\documentclass[10pt,twocolumn,letterpaper]{article}

\usepackage[applications]{wacv}      
\usepackage{comment}
\usepackage{wrapfig}
\usepackage{array}
\usepackage{float}
\usepackage{makecell} 
\usepackage{booktabs} 
\usepackage{multicol}
\usepackage{multirow}

\newcommand{\pname}{\text{RadPose}\xspace}

\definecolor{wacvblue}{rgb}{0.21,0.49,0.74}
\usepackage[pagebackref,breaklinks,colorlinks,allcolors=wacvblue]{hyperref}

\def\wacvPaperID{236} 
\def\confName{WACV}
\def\confYear{2027}

\title{\pname: ISAC-based RF 3D Human Pose Estimation}

\author{Geo Jie Zhou \quad Kunzhe Song \quad Ruxin Lin \quad Huacheng Zeng\\
Department of Computer Science and Engineering, Michigan State University\\
{\tt\small \{geozhou, songkunz, linruxin, hzeng\}@msu.edu}}

\begin{document}
\maketitle
\begin{abstract}
 
Human pose estimation under adverse lighting and occlusion conditions remains challenging for camera- and LiDAR-based systems. Although mmWave radars provide robustness to illumination and occlusion, they rely on dedicated hardware and licensed spectrum, limiting scalability.
In this paper, we present \pname, a learning-based monostatic integrated sensing and communication (ISAC) framework that enables mmWave communication devices to perform 3D human pose estimation. Unlike specialized radars, \pname leverages monostatic full-duplex operation to extract channel impulse responses (CIRs) directly from standard OFDM communication signals, generating 3D radio point clouds without additional sensing hardware.
To address the sparsity and noise of RF observations, \pname employs a cross-frame transformer-based pose inference network with semantic-guided spatial attention and target-driven temporal attention for robust spatio-temporal aggregation. It further adopts a joint-aware inverse kinematics refinement module based on graph convolution to enforce skeletal topology, bone-length consistency, and anatomical plausibility.
Extensive experiments demonstrate \pname achieves competitive accuracy compared to state-of-the-art radar-based approaches while maintaining robust performance in unseen and occluded scenarios. The dataset and source code are available at \href{https://github.com/Geo-Chou/RadPose}{https://github.com/Geo-Chou/RadPose}.

\end{abstract}
    
\vspace{-0.2in}
\section{Introduction}
\label{sec:intro}


Human pose estimation is a fundamental problem in computer vision
~\cite{zheng2023deep}. 
Accurate 3D pose estimation allows machines to interpret human actions, infer intent, and interact safely in dynamic environments. Existing solutions predominantly rely on vision sensors such as cameras and LiDAR, which have driven remarkable progress in recent years~\cite{sun2019deep, huang2025posemamba, ren2024livehps}. However, both camera and LiDAR sensors  depend on light propagation and reflection, making them vulnerable to poor illumination, adverse weather, and severe occlusions in cluttered scenes~\cite{lee2023human, crescitelli2020poison, cheng2019occlusion, ren2024livehps}. 

Millimeter-wave (mmWave) radar has emerged as a compelling alternative for human pose estimation, with prior work demonstrating remarkable performance~\cite{choi2025mvdoppler, xue2021mmmesh, hu2024mmpose}.  
Unlike camera and LiDAR, radar operates using radio-frequency (RF) waves, enabling robustness to lighting conditions and partial penetration through visual occlusions such as clothing, smoke, or lightweight obstacles. This property makes radar attractive for all-weather perception. Nevertheless, conventional radar systems typically require multi-GHz spectrum bandwidth to achieve the desired ranging resolution for pose reconstruction. Such bandwidth either demands dedicated licensed spectrum from regulatory bodies like the Federal Communications Commission (FCC) or must share spectrum with incumbent wireless systems, raising concerns about mutual interference with Wi-Fi or cellular networks. Additionally, deploying standalone radar hardware increases system cost and limits scalability for widespread adoption~\cite{kim2025multi}.

\begin{table}[t]
\centering
\vspace{-0.1in}
\caption{Comparison of \pname and other sensing modalities.}
\vspace{-0.1in}
\resizebox{\linewidth}{!}{
\begin{tabular}{lcccccccc}
\toprule
\textbf{Tech.} &
\makecell{\textbf{Medium}} &
\makecell{\textbf{Waveform}} &
\makecell{\textbf{Occlusion}\\ \textbf{Resilience}} &
\makecell{\textbf{Spectrum}\\ \textbf{License}} &
\makecell{\textbf{Hardware}} &
\makecell{\textbf{Power}\\ \textbf{Consump.}} &
\makecell{\textbf{Cost}} &
\makecell{\textbf{\# of Devices}\\ \textbf{Needed}} \\
\midrule
Camera & Light & --- & Poor & --- & Dedicated & Low & Low & 1 \\
LiDAR & Laser & Pulsed & Poor & --- & Dedicated & High & High & 1 \\
Radar & Radio & FMCW & Good & Required & Dedicated & Med & Med & 1 \\
\multirow{2}{*}{ISAC} & WiFi CSI & OFDM & Good & Not required & Reuse & $\sim${Zero} & $\sim$Zero & $\ge2$ \\
& \textbf{\pname} & \textbf{OFDM} & \textbf{Good} & \textbf{Not required} & \textbf{Reuse} & $\sim$\textbf{Zero} & $\sim$\textbf{Zero} & \textbf{1} \\
\bottomrule
\end{tabular}
}
\label{tab:isac_comparison}
\vspace{-0.2in}
\end{table}

In this paper, we introduce \pname, a learning-based integrated sensing and communication (ISAC) framework that enables individual mmWave communication devices to estimate human pose within their field of view. Table~\ref{tab:isac_comparison} provides a comparison between \pname and other sensing modalities. By embedding sensing capabilities directly into communication nodes, \pname eliminates the need for dedicated radar spectrum licenses, specialized sensing hardware, or separate radio emissions solely for sensing purposes. Instead, it  reuses OFDM communication signals for environmental perception, avoiding harmful interference to incumbent systems while enabling ubiquitous, low-cost deployment. This integration transforms widely-deployed wireless communication infrastructure into a distributed sensing network, opening a pathway toward scalable human pose estimation in scenarios where vision sensors fail or privacy is a constraint. 


RadPose comprises three key components. The first component is a reliable sensing data acquisition mechanism. Most existing ISAC-based sensing systems adopt a \emph{bistatic} configuration, where physically separated transmitter and receiver nodes collect channel state information (CSI). In such settings, dynamic variations in the transmitter-receiver distance, pose, and orientation introduce uncertainties in the measured CSI, limiting generalization across environments. More importantly, bistatic CSI measurements lack ranging information, making object distance estimation impossible. To address these challenges, \pname enables \emph{monostatic} full-duplex operation on mmWave communication devices, leveraging the strong isolation between transmit and receive antenna arrays. The resulting channel impulse response (CIR) measurements provide both ranging and angular information, which are further processed into 3D point clouds suitable for human pose estimation.

The second component is a learning-based pose estimation network that maps radio point clouds to a structured human skeleton. RF-derived point clouds are inherently sparse, noisy, and affected by near-far signal attenuation. To overcome these challenges, \pname introduces Physics-Aware Sparse Convolution (PASC), Semantic-Guided Spatial Attention (SGSA), and Target-Driven Temporal Attention (TDTA). PASC aggregates information from three dimensions with different physical attributes. SGSA leverages multi-scale semantic features from a feature pyramid network along with learnable queries to suppress background noise and enhance human-centric spatial representations. Further, TDTA adopts an asymmetric frame compression strategy with learnable target-frame queries to aggregate temporal context efficiently while mitigating cross-frame feature confusion.

The third component is an inverse kinematics (IK) refinement module that enforces kinematic consistency and temporal coherence of the predicted human skeleton. Due to the low spatial resolution and high noise level of RF data, predicted poses may exhibit structural collapse or joint ambiguity, where neighboring joints become indistinguishable. To address this issue, \pname adopts joint-aware inverse kinematics refinement via a graph convolutional network (GCN) operating on joint-centric embeddings. This module refines the pose estimates by enforcing physical constraints and human body priors, ensuring anatomically plausible and temporally stable pose reconstruction.

We built a prototype of \pname and evaluated it in realistic scenarios. It achieves competitive pose estimation accuracy compared with SOTA radar-based methods, with average estimation errors of 16.2\,cm in seen environments and 18.8\,cm in unseen environments. Moreover, \pname maintains stable performance under occlusion, in stark contrast to camera- and LiDAR-based systems.

The contributions of this work are as follows:
\begin{itemize}
\item [\textbullet] We present a reliable RF sensing data acquisition method on individual mmWave communication devices, enabling monostatic sensing without dedicated radar hardware or licensed spectrum.  

\item [\textbullet] We address the low spatial resolution of RF sensing with a learning-based pose estimation network that combines semantic-guided spatial attention and target-driven temporal attention, while enforcing kinematic consistency via graph-based refinement.  

\item [\textbullet] Extensive experiments demonstrate \pname's high pose estimation accuracy in both seen and unseen environments, as well as robust performance under occlusions.
\end{itemize}

\section{Related Work}
 
\subsection{Camera- and LiDAR-based Pose Estimation}

Camera-based 3D human pose estimation has been extensively studied. Existing methods can be classified into two main categories. The first directly predicts 3D joint coordinates from images \cite{chun2023learnable, iskakov2019learnable, reddy2021tessetrack}. The second follows a two-stage pipeline, where 2D keypoints are first estimated and then lifted to 3D space \cite{mehraban2024motionagformer, sun2019deep, yang2023effective}. Temporal networks \cite{cheng20203d, pavllo20193d} and graph networks \cite{cai2019exploiting, ci2019optimizing, wang2020motion} have been developed to model time sequences and relationships between joints, so as to improve the estimation accuracy. Recently, transformer-based methods \cite{zheng20213d, einfalt2023uplift, li2022exploiting, shan2022p, shan2023diffusion, tang20233d, liu2025tcpformer, zheng2025hipart, huang2025posemamba} have become the mainstream technique for 3D pose estimation.
%
Although camera-based models have demonstrated remarkable performance, they do not perform well in poor-lighting and occluded scenarios.



Compared to camera-based methods, LiDAR-based human pose estimation can accurately localize human joints regardless of illumination. Existing approaches generally fall into two categories: pure LiDAR estimation and multi-modal estimation. Pure LiDAR methods employ point-cloud feature extractors such as PointNet and PointNet++ to learn geometric representations \cite{Ge_2018_ECCV, li2022lidarcap}, with subsequent works enhancing generalization through techniques like knowledge distillation and motion coherence \cite{ren2024livehps, ren2024livehps++}, transformer-based global modeling \cite{Ye2023LPFormerLP, An2024PretrainingAD}, and context-aware modules to better separate human points from noise \cite{ren2024livehps++, Zhang2024Neighborhood}. Multi-modal methods instead fuse LiDAR with camera data to exploit complementary strengths, as pioneered by HPERL \cite{furst2021hperl}, which mitigates RGB depth ambiguity and LiDAR sparsity; later studies extended this strategy to weak supervision and multi-person settings \cite{cong2023weakly, Zheng2021Multimodal3H, Pan2025LiCamPoseCM}. Nevertheless, LiDAR-based approaches still struggle with occlusion, and the high cost of LiDAR sensors limits their widespread adoption.

\subsection{RF-based Pose Estimation}
\noindent\textbf{MmWave Radar Pose Estimation.} 
MmWave radar has demonstrated its abilities to estimate human 3D pose
\cite{zhao2018rf, zheng2022recovering, xiao2023novel, li2024mmskeleton, geng2022densepose, yan2024person}. 
Prior work explored diverse radar data representations including range-Doppler spectra \cite{ding2021radar}, radar point clouds \cite{sengupta2020mm, li2024mmskeleton, xie2024mikey, wu2024mmhpe, wang2025points, Song_2026_CVPR}, and multi-view heatmap encodings \cite{choi2025mvdoppler, xie2023rpm, wu2024mmhpe}. 
To address the spatial sparsity and high noise level of radar data, these methods often employ modules such as multi-frame fusion and spatio-temporal attention \cite{ adhikari2023argosleep, xie2023rpm, hu2024mmpose, wu2024mmhpe, xie2025mmdiffusion}. Beyond pose skeleton estimation, several works focus on reconstructing full human body meshes \cite{xue2021mmmesh, xue2022m4esh, xue2023towards, yang2024mmbat}. Other studies investigate extreme sensing conditions, such as through-wall or around-corner scenarios \cite{zheng2022recovering, lv2025mmwave}. 
To enhance the structural correctness, prior work has incorporated various joint refinement strategies like kinematic priors into radar-based pose reconstruction \cite{ding2021radar, hu2024mmpose, xue2022m4esh, zhu2025probradarm3f, wu2025robust}.
Despite the promise of radar sensing, its deployment requires dedicated hardware and licensed spectrum. This limitation motivates us to explore fine-grained RF sensing using existing communication infrastructure.


\noindent
\textbf{WiFi CSI-based Pose Estimation.}
WiFi CSI-based sensing is a representative ISAC approach that repurposes commodity communication devices for sensing. It has the potential for large-scale deployment because it requires neither new hardware nor licensed spectrum.
Early studies demonstrated the feasibility of recovering human poses from WiFi signals using conventional learning-based models \cite{ren2021winect, geng2022densepose, ren2022gopose, yang2022metafi, zhou2022perunet, bi2022novel, chen2023seeing, wang2019person}. Subsequent studies employed Transformer-based architectures to improve feature representation, expand sensing coverage, and support multi-person pose estimation \cite{wang2023wilink, zhou2023metafi++, zhou2024adapose, yin2024powerskel, yan2024person}.
However, WiFi CSI-based sensing often generalizes poorly to unseen environments. More recent studies, such as DT-Pose, have therefore focused on cross-environment generalization \cite{chen2026dt, hao2025latent, wang2025efficient, wang2025efficientwipose, gian2025multi, chen2026graph, qu2026multiformer}. Despite these efforts, fundamental generalization challenges remain. This is because CSI measurements obtained in \textit{bistatic} configurations lack a consistent physical interpretation owing to Tx/Rx hardware impairments and uncertainties in the Tx/Rx geometry \cite{song2024siwis, 11421930}.

\noindent
\textbf{Positioning of \pname.}
\pname investigates \emph{monostatic ISAC} using a single mmWave communication device for sensing. Compared with radar-based approaches, it requires neither dedicated sensing hardware nor access to licensed spectrum. Compared with WiFi CSI-based methods, its CSI measurements provide physically interpretable features (e.g., ranging, micro-Doppler, and angle of arrival), enabling better generalization to unseen environments. We compare \pname with other sensing modalities in Table~\ref{tab:isac_comparison}.

\section{Monostatic ISAC for Data Acquisition}

\begin{figure}[t]
\centering
  \includegraphics[width=0.75\linewidth]{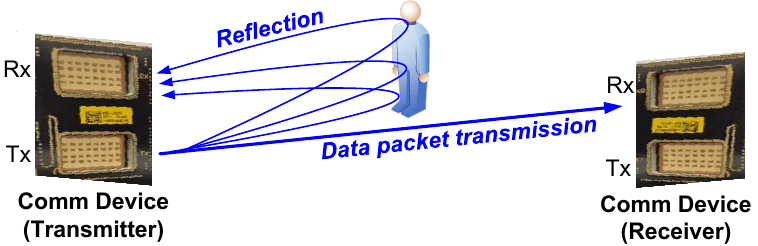}
  \vspace{-0.1in}
  \caption{A mmWave communication device operates in full-duplex mode. It is transmitting OFDM signal for data packet transmission; at the same time, its receiver demodulates the OFDM symbols to estimate CIR, which is used for sensing applications.}
  \label{fig:cir1}
  \vspace{-0.1in}
\end{figure}

MmWave communications have been widely adopted in modern wireless systems, including Wi-Fi (802.11ad) standards and 5G FR2 networks. As illustrated in Figure~\ref{fig:cir1}, when a mmWave Wi-Fi or 5G device transmits data packets, its OFDM signals also illuminate the surrounding environment. If the device can simultaneously transmit and receive, its receiver can capture the backscattered signals from the surrounding objects and estimate the time-domain channel impulse response (CIR). In such a monostatic setting, these CIR measurements enable precise object ranging, allowing the device to estimate distances to surrounding objects in a manner analogous to monostatic radar.
In real systems, mmWave communication devices indeed can operate in full-duplex mode due to their highly directional phased-array antennas and short carrier wavelengths, which together provide sufficient transmitter-receiver isolation.


\subsection{CIR Measurement and Interpretation}

In practice, most mmWave devices use OFDM waveforms for data packet transmission.
When a device is transmitting OFDM signals for data transmission, it can receive the backscattered OFDM signals if operating in full-duplex mode (see the transmitter in Figure~\ref{fig:cir1}). 
The received OFDM signals can be used to estimate the channel response, constituting the input data for downstream sensing tasks.

Denote $K$ as the number of subcarriers in an OFDM symbol. Denote $Y_k$ as the received data symbol and $X_k$ as the transmitted data symbol on OFDM subcarrier $k$.
Then, the channel coefficient of OFDM subcarrier $k$ can be computed by: $H_k = Y_k/X_k$. 
With the channel coefficient of all subcarriers, the CIR can be calculated by:
$
\mathbf{h} = [h_0, h_1, \dots, h_{K-1}] = \operatorname{IFFT}\big([H_{0}, H_{1}, \ldots, H_{K-1}]\big)
$.

When a device operates in full-duplex mode, its transmitter and receiver are co-located on the same device and thus perfectly synchronized in both timing and clock domains. 
As a result, the CIR measurement can be directly interpreted to object ranging information. Specifically, a large magnitude of CIR coefficient $h_k$ indicates a high probability of an object present at a distance of $\frac{kc}{2W}$, where $c$ is light speed and $W$ is OFDM signal bandwidth. 
This ranging capability fundamentally differs \pname from WiFi CSI-based sensing methods.

\subsection{Radio Point Cloud Generation}
\label{sec:radio_point}
\begin{figure}
    \centering
    \includegraphics[width=0.35\linewidth]{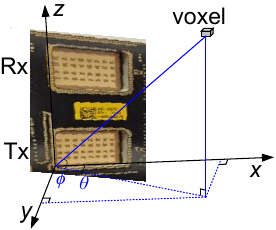}
    \vspace{-0.1in}
    \caption{Angular estimation on a mmWave  device. 
    }
    \label{fig:cir2}
    \vspace{-0.15in}
\end{figure}

Commercial mmWave communication devices are widely equipped with phased-array antennas for signal beamforming. 
The phased-array antennas can be leveraged for angular estimation based on the CIR measurement, generating the point cloud for downstream tasks.
Consider a voxel in the 3D space as shown in Figure~\ref{fig:cir2}. 
Denote $\theta$ as its azimuth angle and $\phi$ as its elevation angle.
The direction of this voxel can then be written as:
$\vec{u}(\theta, \phi) = 
[\cos(\theta)\cos(\phi), 
~\cos(\theta)\sin(\phi), 
~\sin(\phi)]$.

Denote $\mathcal{A}_{tx}$ as the set of Tx antennas and $\mathcal{A}_{rx}$ as the set of Rx antennas. 
Denote $\vec{\tilde{p}}_i = (\tilde{x}_i, 0, \tilde{z}_i)$ as the coordinate of Tx antenna $i \in \mathcal{A}_{tx}$ and 
$\vec{\bar{p}}_j = (\bar{x}_j, 0, \bar{z}_j)$ as the coordinate of Rx antenna $j \in \mathcal{A}_{rx}$. 
When antenna $i$ transmits and antenna $j$ receives signal, the additional signal travel distance in the direction of $(\theta, \phi)$ can be computed by: 
$\langle\vec{u}(\theta, \phi), ~ \vec{\tilde{p}}_i + \vec{\bar{p}}_j\rangle$. 
Therefore, the complex weight associated with the antenna pair $(i, j)$ can be written as:
\begin{equation}
    \alpha_{i,j}(\theta, \phi) = \exp\big(-\mathrm{j}\frac{2\pi}{\lambda} 
    \langle\vec{u}(\theta, \phi), ~ \vec{\tilde{p}}_i + \vec{\bar{p}}_j\rangle\big),
    \label{eq:weights}
\end{equation}
where $\lambda$ is wavelength of carrier signal and the first $j$ is the imaginary unit.

Denote
$\mathbf{h}_{i,j} = [h_{i,j,0}, h_{i,j,1}, \dots, h_{i,j, K-1}]$ as the CIR measurement when \pname uses antenna $i$ for signal transmission and antenna $j$ for signal reception.
Then, the reflected signal from voxel $(k, \theta, \phi)$ can be estimated by:
\begin{equation}
    s(k, \theta, \phi) 
    = 
    \sum_{i \in \mathcal{A}_{tx}}  \sum_{j \in \mathcal{A}_{rx}} \alpha_{i,j}(\theta, \phi) \cdot h_{i,j,k} \,,  
    \label{eq:beamformed_csi}
\end{equation}
where $k$ represents the voxel distance (\textit{i.e.}, $r = \frac{kc}{2W}$).
The magnitude of $s(k, \theta, \phi)$ represents the likelihood of an object at this voxel position, while the temporal phase variation of $s(k, \theta, \phi)$ reveals the object's velocity, if present.

For radio point cloud generation, let $\mathcal{K}$ be the set of range indices, $\Theta$ the set of azimuth angles, and $\Phi$ the set of elevation angles. A radio point cloud frame can be    obtained by evaluating Eq.~\eqref{eq:beamformed_csi} over these domains, which we denote as
\begin{equation}
\mathbf{S} = \big\{ s(k, \theta, \phi) \;|\; k \in \mathcal{K},\ \theta \in \Theta,\ \phi \in \Phi \big\}.
\label{eq:radio_cloud}
\end{equation}

Appendix~A provides more details on the radio point cloud generation and illustration.





\section{Method}

\begin{figure*}[t]
    \centering
    \includegraphics[width=0.87\textwidth]{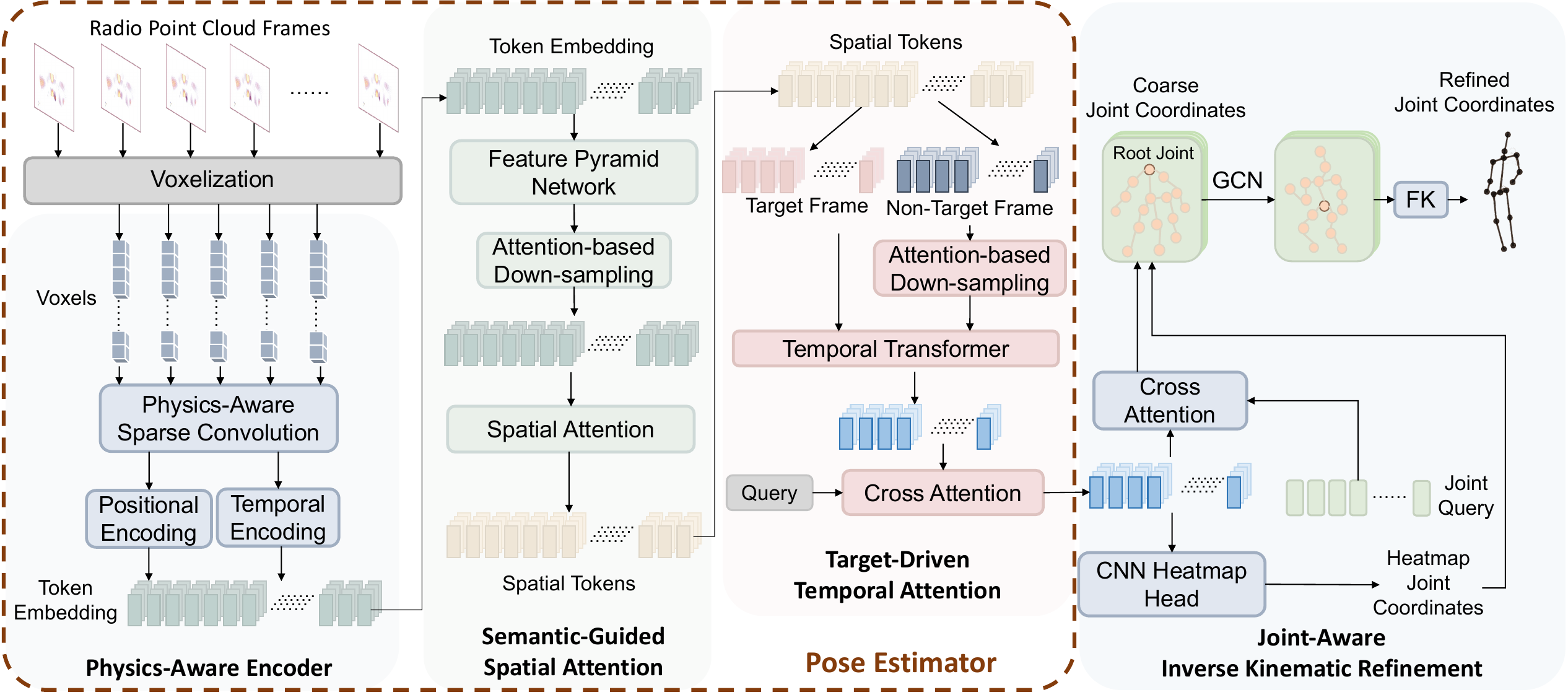}
    \vspace{-0.1in}
    \caption{Overview of \pname. Physics-aware encoder and spatio-temporal attention aggregate spatio-temporal features from radio point clouds, and extract target frame's feature for joint estimation. Joint-aware inverse kinematic refinement then estimates initial heatmaps and applies structural constraints to generate the refined 3D coordinates.}
    \label{fig:software-overview}
    \vspace{-0.1in}
\end{figure*}

\pname takes consecutive radio point cloud frames as input to estimate the 3D human pose.
Figure~\ref{fig:software-overview} illustrates the network pipeline. It consists of two main components: (i) a pose estimator and (ii) an inverse kinematics module.
The pose estimator first voxelizes the noisy radio point clouds, followed by a physics-aware encoder and spatio-temporal attention modules. A heatmap prediction head then produces joint heatmaps by learning discriminative spatial and temporal representations from the radio point cloud sequences.
The inverse kinematics module employs a GCN, where the predicted heatmaps serve as node features. By explicitly modeling human kinematic constraints, this module refines the pose skeleton generated by the pose estimator to ensure anatomically consistent predictions.

\subsection{Pose Estimator}
\label{sec4-1:model}

\textbf{Voxelization.}
The radio point clouds in Eq.~\eqref{eq:radio_cloud} are represented in spherical coordinates. 
Within the spherical region of interest, we partition the space into voxel grids by uniformly discretizing the radial distance, azimuth angle, and elevation angle.
For each radio frame, we first filter out points whose amplitude values are below a threshold, which is empirically set. Points with low intensity are likely caused by circuit noise or multipath reflections, and thus do not carry object information.
The remaining radio points are then discretized into a 3D spherical voxel grid. The resulting set of voxels at frame $t$ is defined as:
$
\mathbf{V}_t = \{ (\mathbf{p}_i, \log(c_i), a_i, \phi_i) \mid c_i > 0, i \in \mathcal{I} \}
$,
where $t$ denotes the frame index, $\mathcal{I}$ is the set of valid voxel indices, $\mathbf{p}_i$ is the center coordinate of voxel $i$, $c_i$ is the number of radio points within voxel $i$, $a_i$ is the average amplitude of the points within voxel $i$, and $\phi_i$ is the average phase of the points within voxel $i$.
In practice, the majority of voxels are empty.
We thus remove empty voxels to reduce the data dimensionality for subsequent processing.

\noindent
\textbf{Physics-Aware Encoder.}
In the spherical coordinate system, the spatial extent covered by a voxel increases with radial distance (\textit{i.e.}, the near-far effect).
Conventional sparse convolution treats each dimension of voxels as isotropic, which is appropriate in Cartesian space where voxel extents are uniform. However, in spherical space, dimensions split into radial and angular ones, each exhibiting distinct physical characteristics. Using conventional sparse convolution tends to cause feature entanglement and suboptimal spatial modeling. 

 To address this, \pname performs \textit{anisotropic} convolution on each individual radio frame over the three branches for dimension-specific voxel feature extraction. Then, it performs physics-aware aggregation through a linear combination of the resulting embeddings, followed by positional encoding along the spatial and temporal dimensions. We denote \(\mathbf{X}_t\) as the output encoded token from this module, corresponding to the voxelized radio frame $\mathbf{V}_t$. Details of the design can be found in Appendix~B.



\noindent
\textbf{Semantic-Guided Spatial Attention.}
Due to the sparsity and high noise level of RF-derived data, a large number of background voxels overwhelm the informative human voxels. In addition, the scale diversity of human movements is challenging for Transformers to model. As a result, spatial attention alone is insufficient.
To address this, \pname first employs a Feature Pyramid Network (FPN) to extract multiscale features, which provide more detailed semantics indicating region of interest for transformer and remain robust for human detection during movements. 
Then, it adopts learnable queries to provide additional semantics for token downsampling.
This operation enhances useful tokens for pose estimation and reduces transformer memory consumption. 
The spatial features of downsampled tokens, which we denote as \(\mathbf{X}^\mathrm{s}_t\), are extracted by spatial self-attention. 

\noindent
\textbf{Target-Driven Temporal Attention.}
Temporal attention across multiple consecutive frames is essential for improving estimation accuracy. 
Accordingly, \pname leverages $T$ consecutive radio frames to estimate one pose.
Conventional cross-frame attention treats all $T$ frames equally, resulting in a high-dimensional token space that increases computational complexity for the network. 
To address this issue, \pname designates one of the $T$ frames as the target frame and the others as non-target frames. 
Full tokens are retained only for the target frame, while tokens from non-target frames are compressed into high-semantic tokens. 
This design reduces background noise from non-target frames while still providing complementary semantic cues from neighboring frames, enabling more focused temporal modeling.

Additionally, directly decoding pose from the aggregated feature tends to cause information confusion.
To avoid this, \pname employs a target frame decoding strategy by introducing learnable target-frame queries.
The learnable queries explicitly retrieve pose-relevant information corresponding to the target frame, ensuring the final representation preserves the frame-specific structure while benefiting from temporal aggregation.
Specifically, the operation is 
\(
\resizebox{0.85\linewidth}{!}{$
\mathbf{X}_{\text{out}} = \mathrm{MHA}\left(\mathbf{Q}_{\text{tar}}, \left[\mathbf{X}_{\mathrm{tar}}^\mathrm{s}, \mathbf{X}_{\mathrm{ntar\downarrow}}^\mathrm{s}\right], \left[\mathbf{X}_{\mathrm{tar}}^\mathrm{s}, \mathbf{X}_{\mathrm{ntar\downarrow}}^\mathrm{s}\right] \right),
$}\)
where \(\mathbf{Q}_{\text{tar}}\) is the target frame query;
\(\mathbf{X}_{\mathrm{ntar}\downarrow}^\mathrm{s}\) is the downsampled feature of \(\mathbf{X}_{\mathrm{ntar}}^\mathrm{s}\),
\(\mathbf{X}_{\mathrm{ntar}}^\mathrm{s} = \{\mathbf{X}_t^\mathrm{s} \;|\; 1 \le t \le T, t \neq \mathrm{tar} \}\).


\noindent
\textbf{Heatmap Head.}
Given the large number of voxels, the pose estimation network tends to have sparse gradients as the loss function may not capture every dimension of the input data.
To address this issue, \pname adopts a 2.5D pose heatmap in the \((r,\theta)\) and \((r,\phi)\) planes as the output of the model, following the principle presented in~\cite{iqbal2018hand}. It provides dense information that can significantly reduce the gradient sparsity. 
To generate the 2.5D pose heatmaps, \pname employs a conv/deconv head to reshape the target-frame tokens $\mathbf{X}_{\mathrm{out}}$ into a spatial feature map. 
The feature map is then progressively upsampled using multiple transposed convolutions to produce the range-azimuth heatmaps \(\mathbf{RA}\) and elevation maps \(\mathbf{E}\). 
For each joint, the coordinate is obtained via soft-argmax on the heatmaps, while the confidence is estimated from the entropy of the heatmap distribution. Details of heatmap head are given in Appendix~C.

\subsection{Joint-Aware Inverse Kinematic Refinement}

    

Poses generated by the above model frequently suffer from inconsistent limb lengths, implausible joint angles, and locally collapsed structures. This occurs because, although the transformer backbone is effective at capturing contextual relationships among spatial and temporal tokens, it does not explicitly encode or enforce human body kinematics. As a result, the predicted joint locations may satisfy feature-level correlations but violate fundamental anatomical constraints.

To address this issue, \pname employs a GCN for inverse kinematic refinement.  
By treating joints as nodes and bones as edges, the GCN explicitly enforces relational consistency among adjacent joints and propagates structural information across the kinematic tree. 

\noindent
\textbf{Joint Embedding.}
Let $\hat{\mathbf{P}}=[\hat{\mathbf{p}}_1,\dots,\hat{\mathbf{p}}_J]$ be the joint coordinates decoded from the heatmap head. For each joint $j$, we construct a joint query by combining a learnable embedding with the decoded joint coordinate. Each joint query attends to the transformer tokens to produce a joint-centric feature:
\vspace{-0.07in}
\begin{equation}
\mathbf{h}_j=\mathrm{MHA}(\mathbf{q}_j,\mathbf{X}_{\text{out}},\mathbf{X}_{\text{out}}),\ \ 
\mathbf{q}_j=\mathbf{e}_j+\mathbf{\tilde{W}}_\mathrm{pos}\hat{\mathbf{p}}_j,
\vspace{-0.07in}
\end{equation}
where $\mathbf{e}_j$ is the learnable embedding of joint $j$, and $\mathbf{\tilde{W}}_\mathrm{pos}$ projects the coordinate into the feature space. 
$\mathbf{h}_j$ encodes pose-relevant information for joint $j$.

\noindent
\textbf{GCN for Inverse Kinematics.}
\pname treats the human skeleton as a kinematic graph. The initial node feature of joint \(j\) is formed by fusing the joint-centric feature \(\mathbf{h}_j\), decoded coordinate \(\mathbf{\hat{p}}_j\), and confidence \(c_j\):
$\mathbf{g}_j^{(0)}=\left[\mathbf{h}_j,\mathbf{\hat{p}}_j,c_j\right]$.
The node features are then propagated on the kinematic graph:
\begin{equation}
\mathbf{G}^{(\ell+1)}=\sigma\!\big(\tilde{\mathbf{A}}\mathbf{G}^{(\ell)}\mathbf{W}^{(\ell)}\big),\qquad \ell=0,\dots,L_g\!-\!1,
\end{equation}
where $\mathbf{G}^{(\ell)}$ stacks all node features at layer $\ell$, $\tilde{\mathbf{A}}\in\mathbb{R}^{J\times J}$ is the normalized confidence-aware adjacency matrix.
$\mathbf{W}^{(\ell)}$ is the learnable weight matrix, $\sigma(\cdot)$ is the activation function.

From the final graph features, \pname predicts a direction vector \(\mathbf{d}_j\), a bone-length scale \(s_j\), and a residual root offset \(\Delta\mathbf{p}_r\) using a model parameterized by \(\psi\):
\begin{equation}
[\mathbf{d}_j,s_j,\Delta\mathbf{p}_r]
=
\psi\!\big(\mathbf{g}_j^{(L_g)}\big).
\end{equation}

\noindent
\textbf{Forward Kinematics.}
In the final step, \pname selects one joint as the root joint for pose construction, Let $r$ be its index.
The refined root joint coordinate is computed as
$
\mathbf{p}_r=\mathbf{\hat{p}}_r+\Delta\mathbf{p}_r
$.
For each non-root joint $j$, its position is reconstructed recursively by forward kinematics:
$
\mathbf{p}_j=\mathbf{p}_{\pi(j)}+(l_j s_j)\mathbf{d}_j
$,
where $j \neq r$; 
$\pi(j)$ denotes the parent joint of $j$;
$l_j$ is the predefined canonical bone length; 
and $s_j$ is the predicted scale factor.
Finally, a small residual correction is added to improve flexibility:
$
\mathbf{p}_j^{\mathrm{ik}}=\mathbf{p}_j+\Delta\mathbf{p}_j$, 
which constitutes the final coordinate of joint $j$.

\subsection{Supervised Training}

\pname employs a two-stage supervised training strategy.
It first trains the pose estimator model  and then trains GCN for the inverse kinematic module.

\noindent
\textbf{Pose Estimator Training.}
The loss function of training the pose estimator is a weighted summation of heatmap loss \(\mathcal{L}_{\mathrm{map}}\), coordinate loss \(\mathcal{L}_\mathrm{xyz} \), bone length loss \(\mathcal{L}_\mathrm{len}\) and joint angle loss \(\mathcal{L}_\mathrm{ang}\), 
where 
\(\mathcal{L}_\mathrm{xyz}\), \(\mathcal{L}_\mathrm{len}\) and \(\mathcal{L}_\mathrm{ang}\) are the smoothL1 losses between predicted value and ground-truth labels. 
In particular, $\mathcal{L}_{\text{map}}$ is the most important loss for this training. It alleviates the sparse gradient issue inherent in direct coordinate regression, as it supervises a spatial distribution rather than a single point estimate. 
It is defined as:
\(
\mathcal{L}_{\text{map}}
= -\frac{1}{|\mathrm{\Omega}|}
\sum_{u \in \mathrm{\Omega}}
\big[
\alpha y_u (1-\hat{y}_u)^{\beta}\log\hat{y}_u 
+ (1-y_u)\hat{y}_u^{\beta}\log(1-\hat{y}_u)
\big],
\)
where $u$ indexes all spatial locations on the predicted heatmap and $\mathrm{\Omega}$ is its complete set, $y_u\in[0,1]$ and $\hat{y}_u\in[0,1]$ denote the ground-truth and predicted heatmap values at location $u$, respectively, and $\alpha$ and $\beta$ are the focusing parameters that balance positive and negative samples and emphasize hard examples.

\noindent
\textbf{GCN Training.}
After training pose estimator, we then train the GCN for the inverse kinematic module. 
During this stage, the GCN is trained with full scale learning rate, while the parameters of pose estimator are trained with a small learning rate to preserve the representations learned by the pose estimator while allowing it to be slightly refined to better cooperate with the inverse kinematics module.
In this stage, the loss of root joint \(\mathcal{L}_\mathrm{root}\) is added to penalize the undesired topology of human pose.

\begin{table*}[htbp]
    \vspace{-0.1in}
    \centering
    \footnotesize
    \caption{Quantitative comparison of RadPose against SOTA RF sensing methods. All MPJPE values are absolute errors without root alignment. [S./U. denotes seen/unseen scenes, P.C. denotes radio point clouds, 
    B.W. denotes Bandwidth, `--' denotes `Not Available'.]}
    \label{tab:comparison_results}
    \vspace{-0.1in}
    \begin{tabular}{lccccccccc}
        \toprule
        Methods &
        Input &
        \makecell{\# of scenes\\ (S./U.)}&
        \# of activities&
        Waveform &
        \makecell{B.W.\\ (GHz)} &
        \makecell{$\Delta R$\\(cm)} &
        \makecell{MPJPE \\ (S./U.)} &
        \makecell{$\text{PCK}_{10}$ \\ (S./U.)} &
        \makecell{MPJPE/$\Delta R$ \\ (S./U.)} \\
        \midrule
        Person-in-WiFi 3D \cite{yan2024person} \!\!\!\!\!\!\!& CSI & 3/1 & 8 & OFDM & -- & -- & 9.2/62.6 & --/-- & --/-- \\
        GraphPose-Fi \cite{chen2026graph}& CSI & 3/1 & 14 & OFDM & -- & -- & 16.1/30.2 & 33.3/2.7 & --/-- \\
        \midrule
        mmMesh \cite{xue2021mmmesh}  & P.C. & 1/1 & 8 & FMCW & 3.9 & 3.8 & 2.2/5.1 & --/-- & 0.58/1.34 \\
        MvDoppler \cite{choi2025mvdoppler}  & Range/Doppler & 1/0 & 1 & FMCW & 1.0 & 15.0 & 9.7/-- & --/-- & 0.64/-- \\
        \midrule
        mmDiff \cite{fan2024diffusion}      & P.C. (Our dataset)& 17/4 & Unconstrained & OFDM & 1.2 & 12.2 & 25.1/61.4 & 16.5/0.5 & 2.06/5.04 \\
        Wave2Body \cite{liang2026wave2body} & P.C. (Our dataset)& 17/4 & Unconstrained & OFDM & 1.2 & 12.2 & 35.5/38.9 & 10.9/10.1 & 2.91/3.18 \\
        \textbf{RadPose}                    & \textbf{P.C. (Our dataset)} &\textbf{17/4} &\textbf{Unconstrained} & \textbf{OFDM} & \textbf{1.2} & \textbf{12.2}& \textbf{16.2/18.8} & \textbf{36.3/29.8} & \textbf{1.33/1.54} \\

        \bottomrule
    \end{tabular}
    \vspace{-0.05in}
\end{table*}

\section{Implementation}

\begin{figure}
    \centering
    \includegraphics[width=0.85\linewidth]{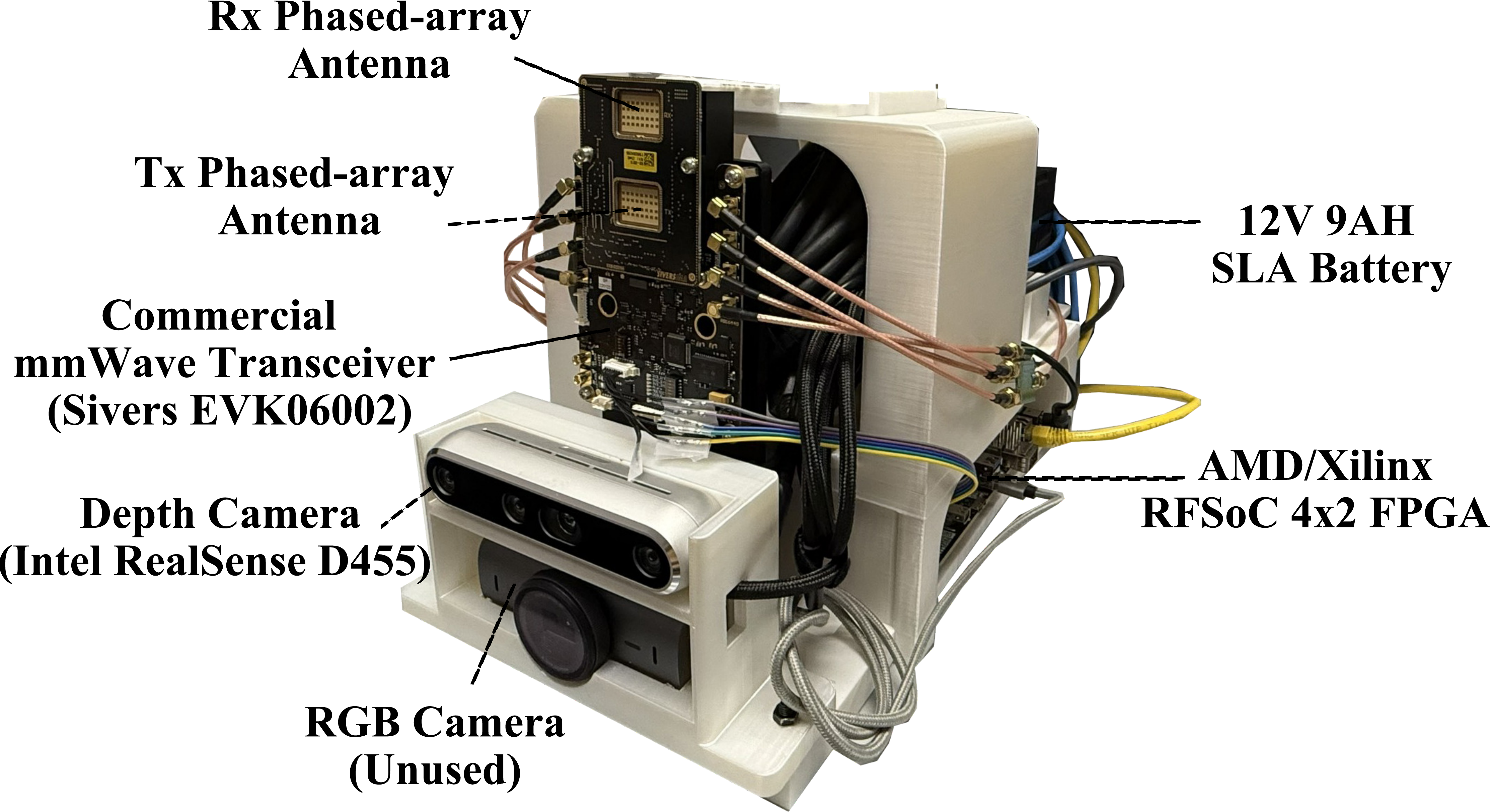}\vspace{-0.05in}
    \caption{Hardware components of \pname.}
    \vspace{-0.1in}
    \label{fig:platform_isac}
\end{figure}

\noindent\textbf{Hardware Setup.}
We built a prototype of \pname, as shown in Figure~\ref{fig:platform_isac}, to evaluate its performance in realistic environments. The hardware integrates two sensing components:
(i) a custom-designed mmWave 5G/Wi-Fi OFDM communication device that supports full-duplex operation for monostatic sensing, and
(ii) a depth camera (Intel RealSense D455) that provides ground-truth pose annotations.

The mmWave device operates at 60\,GHz with a bandwidth of 1.2288\,GHz and $16$ Tx-Rx antenna elements. It offers an sensing range of 6\,m, covering a field of view of $120^\circ$ horizontally and $60^\circ$ vertically. Importantly, the device supports simultaneous data transmission and sensing (see Appendix~E).
The system was built using commercial off-the-shelf RF components.

\noindent\textbf{Network Parameters.}
\pname voxelizes the input into 20,480 voxels, and five consecutive radio frames are used for pose estimation. 
Three physics-aware sparse convolutions encode each voxel into a 512-dimensional token, followed by an attention-based downsampling that reduces the tokens to 2,048. 
For inverse kinematics, the 6th joint is selected as the root, and a three-layer GCN with 512 hidden dimensions is used as the IK head. 
More details are provided in Appendix~D.

\noindent\textbf{Dataset Details.}
{We collected 183,497 radio frames with the ground-truth pose annotations of 11 subjects in 21 scenes. The ground-truth annotations are derived from corresponding depth camera frames with pretrained RTMPose model \cite{jiang2023rtmpose}.
Unlike \cite{choi2025mvdoppler, xue2021mmmesh, chen2026graph}, where subjects are restricted to predefined movements, our dataset covers the subjects without motion sequences, fixed motion amplitudes, or designated activity locations.
Of the 21 scenes, 75\% data of 17 scenes (120,174 frames) are used for model training. The remaining 25\% data (39,400 frames) and holdout 4 scenes (23,923 frames) are reserved for evaluation. We additionally collected 14,209 frames with constrained movements to better evaluate influence of activity types.



\section{Experiments}

\subsection{Main Results}

Two metrics are used for evaluation: 
Mean Per Joint Position Error (MPJPE/cm) and Percentage of Keypoints ($\operatorname{PCK}_\delta$/\%).
$\operatorname{MPJPE} =
\frac{1}{J} \sum_{j=1}^{J} 
\left\| \hat{\mathbf{p}}_j - \mathbf{p}_j \right\|_2$, where $J$ is the total number of joints, $\hat{\mathbf{p}}_j$ and $\mathbf{p}_j$ are estimated and ground-truth positions of joint $j$.
$\operatorname{PCK}_\delta =
\frac{1}{J} \sum_{j=1}^{J}
\mathbb{I}\!\left(
\left\| \hat{\mathbf{p}}_j - \mathbf{p}_j \right\|_2 < \delta
\right)$, 
where $\mathbb{I}(\cdot)$ is the indicator function, \(\delta\) is a joint error distance threshold. Both metrics are based on absolute 3D joint positions without alignment.

\noindent\textbf{Quantitative Results.}
We first evaluate \pname's quantitative performance and its ability to generalize to unseen environments. 
Table~\ref{tab:comparison_results} presents our experimental results. The results show that \pname maintains robust performance on held-out test scenes. Notably, the error of root joint (\textit{i.e.}, the 6th joint) is 12.1~cm, which indicates the precise localization ability of our model.

\noindent\textbf{Comparison with SOTA.}
We compare \pname with SOTA radar-based and WiFi CSI-based pose estimation methods.
MmMesh~\cite{xue2021mmmesh}, Mv-Doppler~\cite{choi2025mvdoppler}, mmDiff~\cite{fan2024diffusion}, and Wave2Body~\cite{liang2026wave2body} are representative FMCW radar-based methods, while Person-in-WiFi 3D~\cite{yan2024person} and GraphPose-Fi~\cite{chen2026graph} are representative WiFi CSI-based methods.
In RF sensing, bandwidth $W$ determines the range resolution, $\Delta R = \frac{c}{2W}$, and fundamentally affects pose estimation accuracy. To account for differences in bandwidth, we use the normalized MPJPE, $\frac{\mathrm{MPJPE}}{\Delta R}$, as the comparison metric. 
Table~\ref{tab:comparison_results} reports the comparison results.

Compared with the WiFi CSI-based methods (Person-in-WiFi and GraphPose-Fi), \pname performs slightly worse in seen environments but substantially better in unseen environments. This improved generalization likely stems from its \emph{monostatic} sensing geometry and physically interpretable signal features, whereas WiFi CSI methods rely more heavily on data-driven feature learning from \emph{bistatic} measurements.

Compared with the radar-based methods (MmMesh and Mv-Doppler), \pname has slightly higher normalized errors: 1.33 versus 0.58 in seen scenes and 1.54 versus 1.34 in unseen scenes. This gap may arise for three reasons. First, MmMesh and Mv-Doppler use dedicated FMCW radars, which are generally more robust to noise than OFDM-based sensing. Second, their evaluation settings are more limited: MmMesh considers two scenes and Mv-Doppler only one, whereas \pname is evaluated in 17 seen and four unseen scenes. Third, they primarily evaluate walking activities, while \pname includes more unconstrained activities. Despite this performance gap, \pname requires neither dedicated sensing hardware nor licensed spectrum and therefore targets different application domains.

To evaluate our network pipeline, we also test mmDiff and Wave2Body on our dataset. As shown in Table~\ref{tab:comparison_results}, \pname substantially outperforms both methods, demonstrating the benefits of PASC, SGSA, TDTA, and joint-aware IK refinement.



\noindent\textbf{Qualitative Results.}
Figure~\ref{fig:qualitative-performance} presents a visual comparison of our pose estimation. 
The first row are scenes used in training, and the second row are unseen scenes. From visualization, most errors come from wrists and ankles, which are difficult for \pname to detect due to their small physical size.
More pose visualization is provided in Appendix F-G, and a demo video is available on the project website.

While most experiments demonstrate accurate pose estimation, we also observe some failure cases in the results. 
Some representative examples are provided in Appendix H.

\begin{figure}[!t]
    \centering
    \vspace{-0.1in}
    \includegraphics[width=0.99\linewidth]{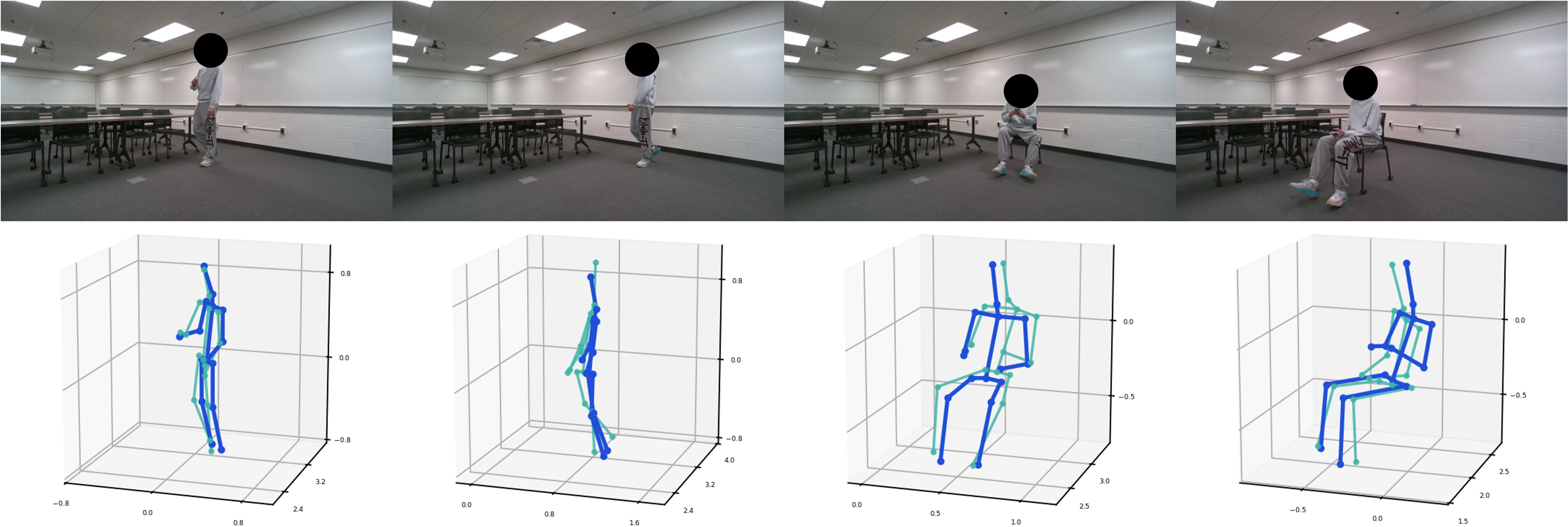}\\
    \includegraphics[width=0.99\linewidth]{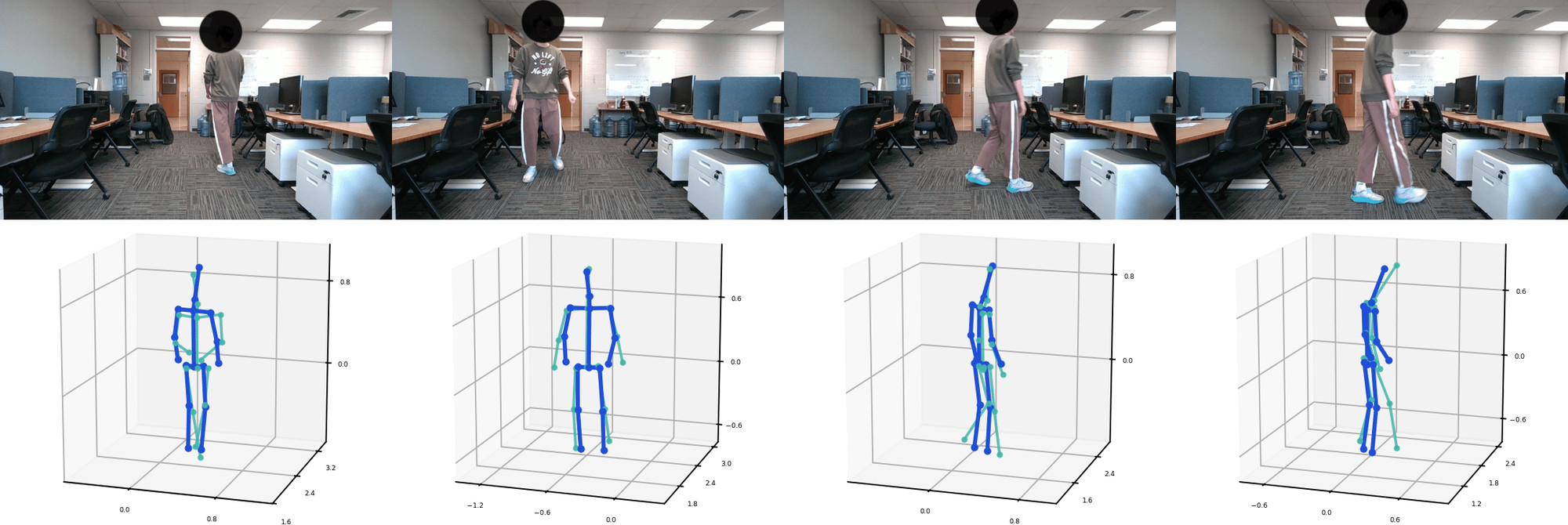}\\
    \vspace{-0.06in}
    \caption{Qualitative performance of \pname in different scenes. The green skeleton represents the ground-truth labels derived from depth camera while the blue skeleton is the results of \pname.}
    \label{fig:qualitative-performance}
    \vspace{-0.1in}
\end{figure}

\noindent\textbf{Impact of Different Activities.}
Table~\ref{tab:activity} reports the performance of \pname across different activities. \pname achieves the highest accuracy for walking, likely because its regular and periodic motion patterns facilitate temporal modeling. Squatting and standing are the most challenging, as compact body configurations bring multiple joints close together, increasing self-occlusion and ambiguity in the reflected signals. Nevertheless, \pname maintains consistent performance on mixed activities, demonstrating its robustness across diverse motions.

\begin{table}[ht]
\vspace{-0.02in}
    \centering
    \footnotesize
    \caption{\pname's performance across different activities.}
    \label{tab:activity}
    \vspace{-0.1in}
    \begin{tabular}{@{}lccccc@{}}
        \toprule
        Activities & Walking & Waving & Sit/Stand & Squat/Stand & Mixed \\
        \midrule
        MPJPE        & 13.17 & 18.72 & 16.07 & 21.56 & 16.21 \\
        $\mathrm{PCK}_{10}$ & 46.9 & 29.7 & 39.1 & 18.4 & 36.3 \\
        \bottomrule
    \end{tabular}
    \vspace{-0.1in}
\end{table}


\noindent\textbf{Robustness to Occlusion.} 
Compared to the camera- and LiDAR-based pose estimation, one salient advantage of \pname is its robustness to occlusions.
To demonstrate this advantage, we conducted experiments using cloth and foam to completely cover the mmWave device's  antennas but leave the depth camera uncovered.
Table~\ref{tab:robustness} reports the experimental results. 
\pname's performance degrades slightly under these occlusions.
In particular, MPJPE remains within 15\,cm and the PCK$_{\mathrm{10}}$ remains above 29\%, indicating \pname maintains a robust performance when the antennas are completely covered.
This reveals that \pname can be used to complement camera and LiDAR in challenging scenarios and its potential for covert surveillance. More occlusion results are given in Appendix H.

\begin{table}[!t]
\centering
\vspace{-0.15in}
\footnotesize
\setlength{\tabcolsep}{2.5pt}
\begin{minipage}[t]{0.48\columnwidth}
\centering
\caption{Occlusion robustness.}
\vspace{-0.07in}
\label{tab:robustness}
\begin{tabular}{lcc}
\toprule
Occlusion & MPJPE& PCK$_{10}$ \\
\midrule
None      & 12.5 & 42.7 \\
Cloth     & 15.0 & 29.2 \\
Styrofoam & 13.4 & 38.2 \\
\bottomrule
\end{tabular}
\vspace{-0.08in}
\end{minipage}
\hfill
\begin{minipage}[t]{0.45\columnwidth}
\centering
\caption{Effectiveness of IK refinement.}
\vspace{-0.1in}
\label{tab:ablation-ik}
\begin{tabular}{lcc}
\toprule
IK & MPJPE & PCK$_{10}$ \\
\midrule
w/o & 17.0 & 14.0 \\
w   & \textbf{16.2} & \textbf{36.3} \\
\bottomrule
\end{tabular}
\vspace{-0.05in}
\end{minipage}
\end{table}

\begin{table}[t]
    \centering
    \footnotesize
    \caption{Subject generalization with and without IK refinement. NE$_{\mathrm{A/L/B}}$ denotes normalized error for arms/legs/body (\%)}
    \label{tab:identity generalization}
    \vspace{-0.1in}
    \setlength{\tabcolsep}{3pt}
    \begin{tabular}{@{}lcccc@{}}
        \toprule
        & \multicolumn{2}{c}{w/o IK} 
        & \multicolumn{2}{c}{w/ IK} \\
        \cmidrule(lr){2-3}\cmidrule(lr){4-5}
        Subjects &
        MPJPE$\downarrow$ &
        NE$_{\mathrm{A/L/B}}$$\downarrow$ &
        MPJPE$\downarrow$ &
        NE$_{\mathrm{A/L/B}}$$\downarrow$ \\
        \midrule
        Seen   & 14.66 & 33.91/18.48/13.21 & 13.91 & 30.80/18.16/12.88 \\
        Unseen & 17.18 & 38.66/21.82/16.19 & 16.30 & 35.41/21.35/15.59 \\
        \bottomrule
    \end{tabular}
    \vspace{-0.1in}
\end{table}

\subsection{Ablation Study}

\noindent\textbf{Joint-Aware IK Refinement.}
The IK refinement is the key component of \pname. 
To evaluate its effectiveness, we measure the pose estimation errors of \pname with and without the IK module. 
Table~\ref{tab:ablation-ik} presents the results.  
The IK module reduces the estimation error from 17.0\,cm to 16.2\,cm, with PCK\(_{10}\) improved 22.3\%. 
This significant improvement can be attributed to the GCN within the IK module, which effectively encodes skeletal nodes and edges to capture human pose geometry.

To evaluate the generalization of our IK refinement module, we measure \pname's performance across subjects. The model was trained on seven-subject dataset with four held out.  Table~\ref{tab:identity generalization} presents the results.
The IK refinement leads to pronounced improvements in estimation accuracy across all subjects.

\noindent\textbf{Other Components.}
More ablation studies of \pname's key components are given in Appendix I.


\section{Conclusion and Discussion}
In this paper, we presented \pname, a monostatic ISAC framework that enables individual mmWave communication devices to perform 3D human pose estimation. The key innovations of \pname include (i) a cross-frame transformer for representing mmWave point clouds and (ii) a novel inverse kinematics method for mmWave sensing that leverages body-topology constraints to improve pose estimation accuracy. Experimental results validate the effectiveness of \pname in realistic environments.

\pname demonstrates the feasibility of using OFDM-based communication signals for fine-grained perception with a single mmWave device. However, the current system has two limitations. First, \pname supports only single-person pose estimation. Future work is needed to extend the learning framework to multi-person scenarios. Appendix K presents a plausible extension of \pname for such scenarios. Second, \pname is not currently compatible with commercial Wi-Fi or 5G communication devices because they do not support full-duplex operation. Integrating \pname into future communication infrastructure will require changes to communication standards, as well as software and firmware upgrades to communication devices. 


\section*{Acknowledgments}
We sincerely thank the reviewers for their insightful comments.
This work is supported in part by the NSF under Award No.~ECCS-2225337 and ECCS-2434001.

\clearpage

{
\small
\bibliographystyle{ieeenat_fullname}
\bibliography{reference}
}

\clearpage
\maketitlesupplementary

In these appendices, we provide additional details on the monostatic ISAC system, the radio point cloud generation, and the network implementation. 
We also present additional experimental results under both normal and challenging scenarios, along with the failure cases of \pname. 
Furthermore, we offer additional ablation studies of \pname and discussions on its extension to multi-person pose estimation.


\section*{Appendix A: Radio Point Cloud Generation}

\begin{figure}[h]
    \centering
    \includegraphics[width=\linewidth]{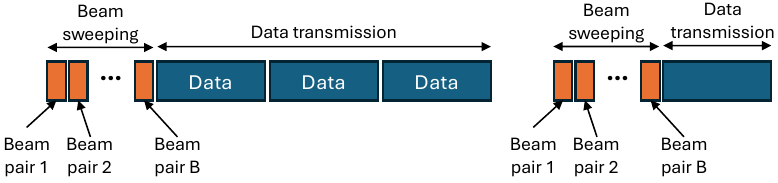}
    \caption{Beam-sweeping protocol in 5G NR systems.}
    \label{fig:beamforming_protocol}
\end{figure}

\noindent
\textbf{Beam Sweeping in mmWave Communication.}
Both 5G FR2 and mmWave Wi-Fi systems periodically perform beam sweeping to establish and maintain directional communication links, as illustrated for 5G NR in Figure~\ref{fig:beamforming_protocol}. During this procedure, the transmitter and receiver sequentially scan predefined codebooks of beam directions by transmitting and measuring reference or training signals. The receiver evaluates and record the signal from each candidate Tx-Rx beam pair, typically using metrics such as received signal strength or signal-to-noise ratio, and selects the pair that provides the best communication link. Meanwhile, the receiver estimates the channel associated with each beam pair. We denote the channel vector for beam pair $b$ by $\tilde{\mathbf{h}}_b$, where $1 \leq b \leq B$. Because user movement and environmental changes can alter the propagation channel, beam sweeping is performed periodically to maintain the quality of an active link.

By leveraging the beam-sweeping protocol, we obtain the per-beam channel matrix
$\tilde{\mathbf{H}} = [\tilde{\mathbf{h}}_1, \tilde{\mathbf{h}}_2, \ldots, \tilde{\mathbf{h}}_B]$,
where $B$ is the total number of Tx--Rx beam pairs. For point-cloud generation, this matrix must be converted into the per-antenna channel matrix
$\mathbf{H} = [\mathbf{h}_{i,j}]$,
where $i$ and $j$ index the Tx and Rx antenna elements, respectively. 
Because the beamforming weights associated with each beam pair are fixed, the relationship between the per-beam and per-antenna channel matrices can be calibrated in advance. Specifically, we recover the per-antenna channel matrix as
$\mathbf{H} = \mathbf{T}\tilde{\mathbf{H}}$,
where $\mathbf{T}$ is a transformation matrix obtained through calibration. The resulting per-antenna channel matrix is then used to generate the radio point cloud, as described in Section \ref{sec:radio_point}.

\noindent\textbf{Point Cloud Illustration.}
Eq.~\eqref{eq:radio_cloud} presents the mathematical representation of a radio point cloud frame. Visualizing all points in a frame produces an overly dense point cloud, making the structure difficult to recognize. For clearer illustration, we define $\mathbf{S}_{>a} = \{|s| \, :\, s \in \mathbf{S}, |s| > a \}$ to display only the strong points in a radio frame, where $\mathbf{S}$ denotes the set of all points in the frame and $a$ represents the point magnitude threshold.

Figure~\ref{fig:radio_frame} shows two examples of radio point cloud frames generated by our mmWave monostatic ISAC platform. The strong points visible in the images originate from both human targets and background objects. Our observations indicate that humans are strong reflectors of mmWave signals, generating prominent points in the radio frame.

In addition to the static images, a demo video is provided on the project website, showing the radio point cloud in real time in dynamic environments.

\begin{figure}[t]
    \centering
    \begin{tabular}{>{\centering\arraybackslash}m{0.3\linewidth}
    >{\centering\arraybackslash}m{0.3\linewidth}
    >{\centering\arraybackslash}m{0.3\linewidth}}
            \textbf{Image} & $\mathbf{S}_{>1}$ &  $\mathbf{S}_{>2}$  \\
            \includegraphics[width=\linewidth]{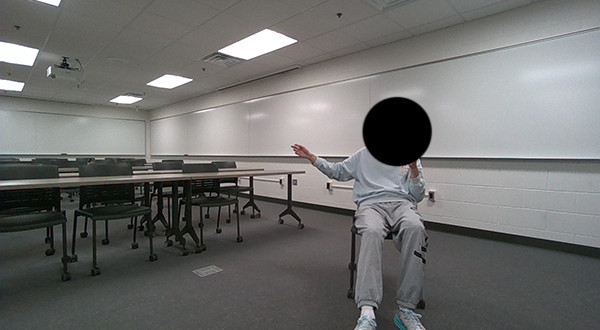} &
            \includegraphics[width=\linewidth]{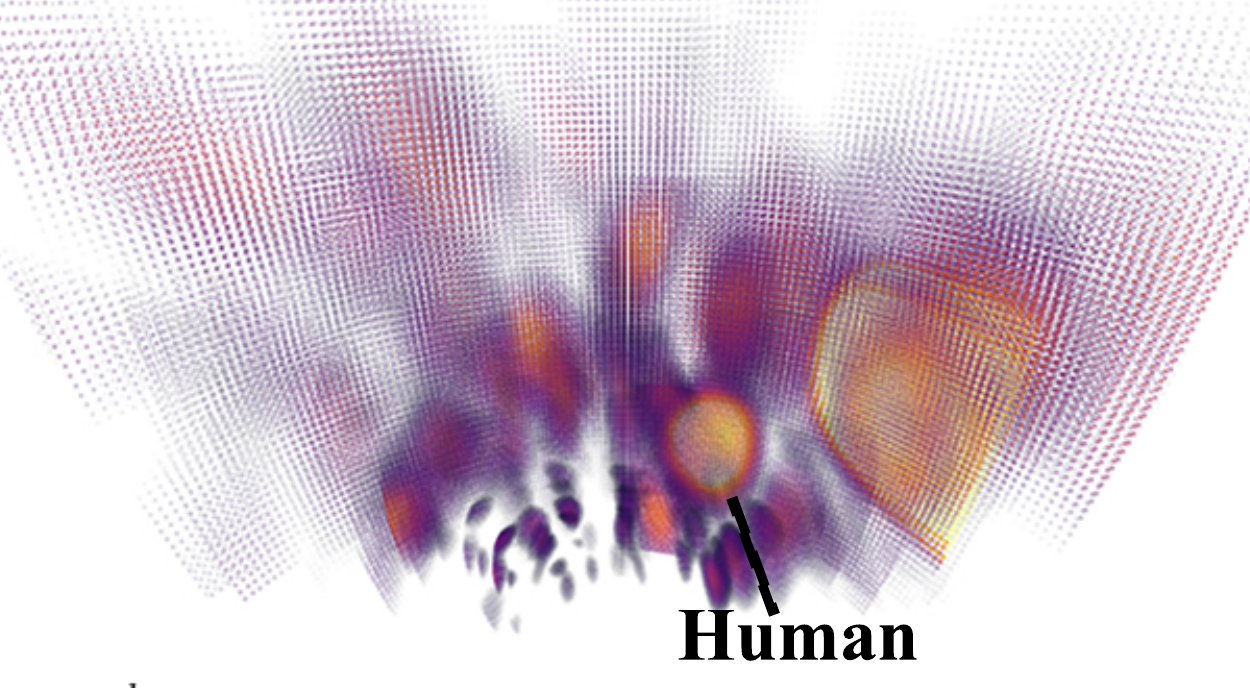} &
            \includegraphics[width=\linewidth]{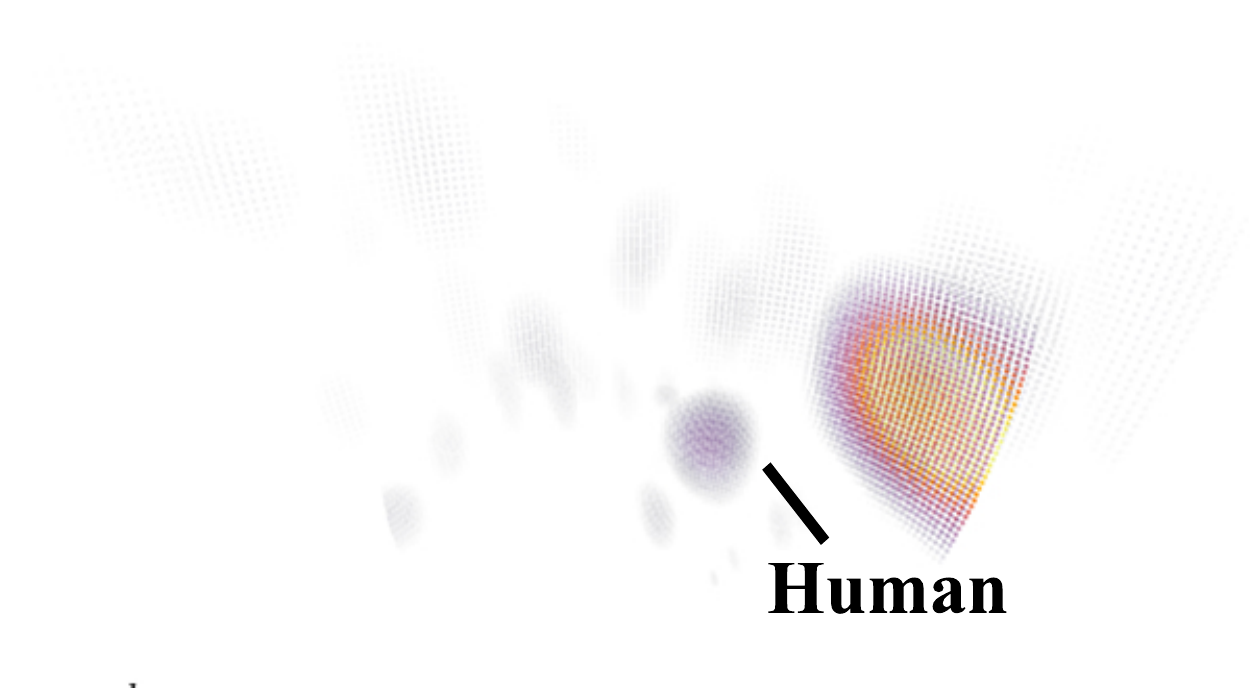}  \\
            \includegraphics[width=\linewidth]{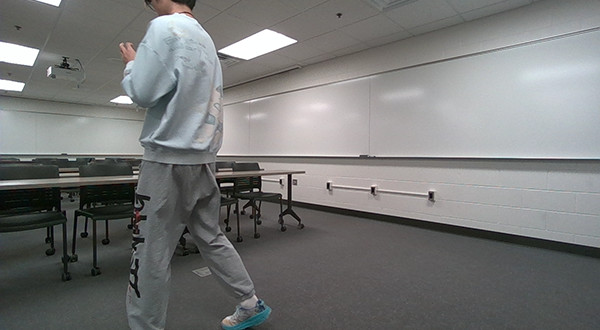} &
            \includegraphics[width=\linewidth]{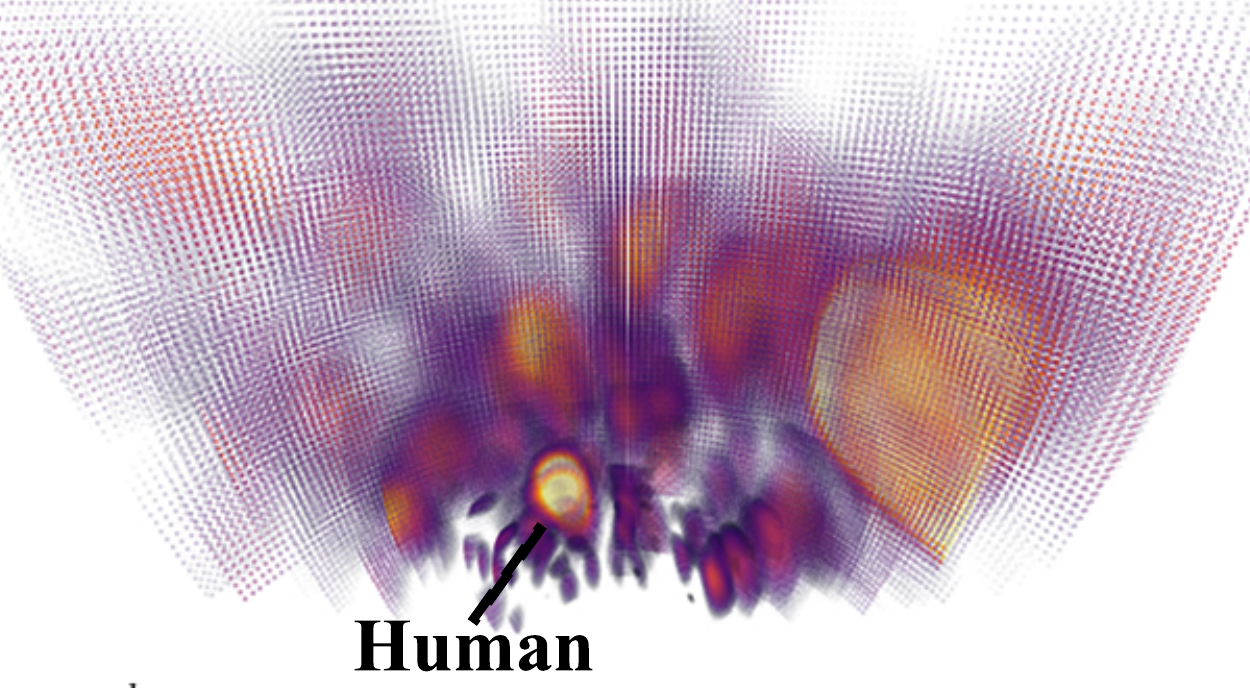} &
            \includegraphics[width=\linewidth]{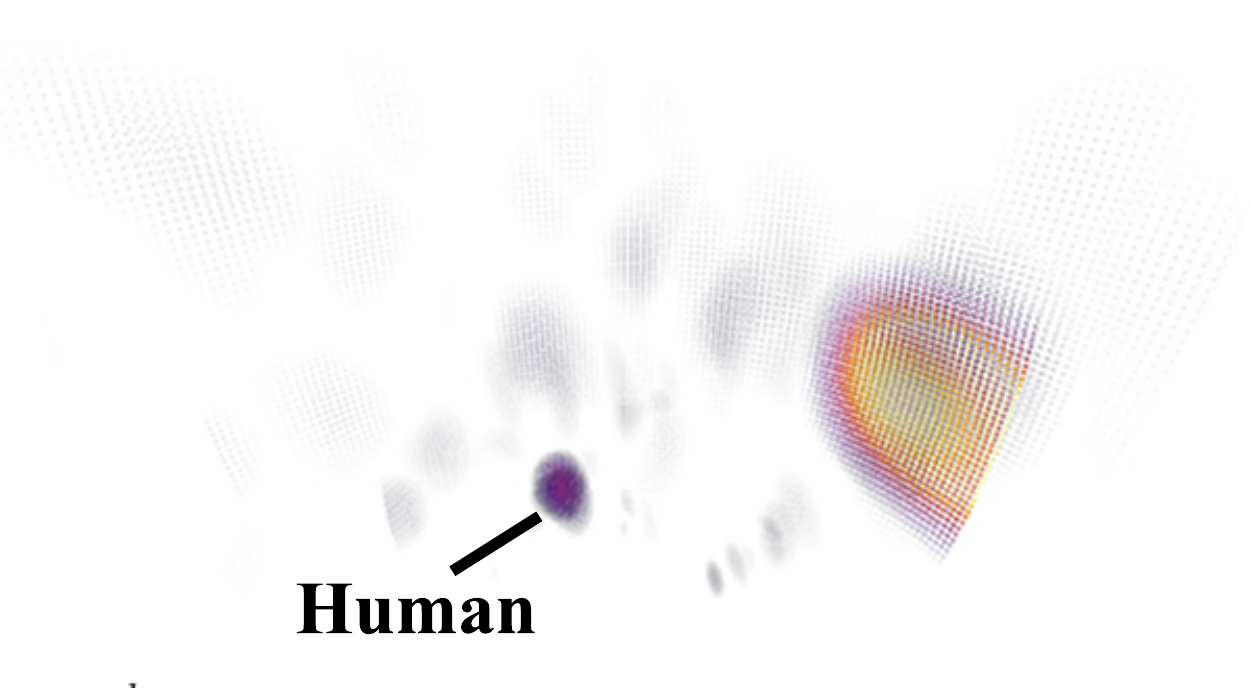} \\
            \end{tabular}
    \caption{Two samples of radio point clouds generated by our mmWave monostatic ISAC platform.}
    \label{fig:radio_frame}
\end{figure}

\begin{figure}[!h]
    \centering
    \begin{minipage}[b]{0.38\linewidth}
        \centering
        \includegraphics[width=\linewidth]{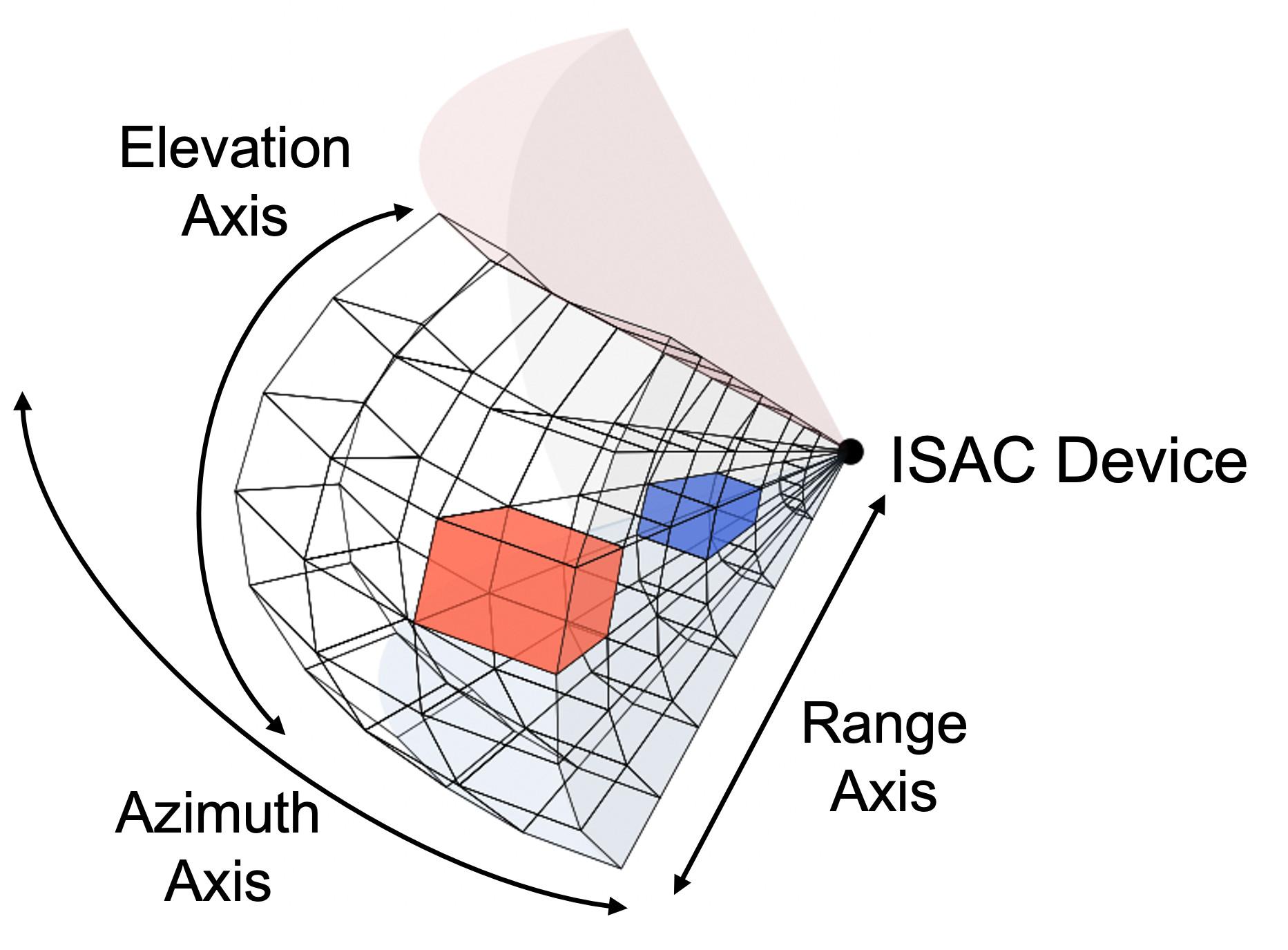}
        \caption{Near-far effect in voxelization.}\label{fig:voxel}
    \end{minipage}
    \hfill    
    \begin{minipage}[b]{0.575\linewidth}
        \centering
        \includegraphics[width=\linewidth]{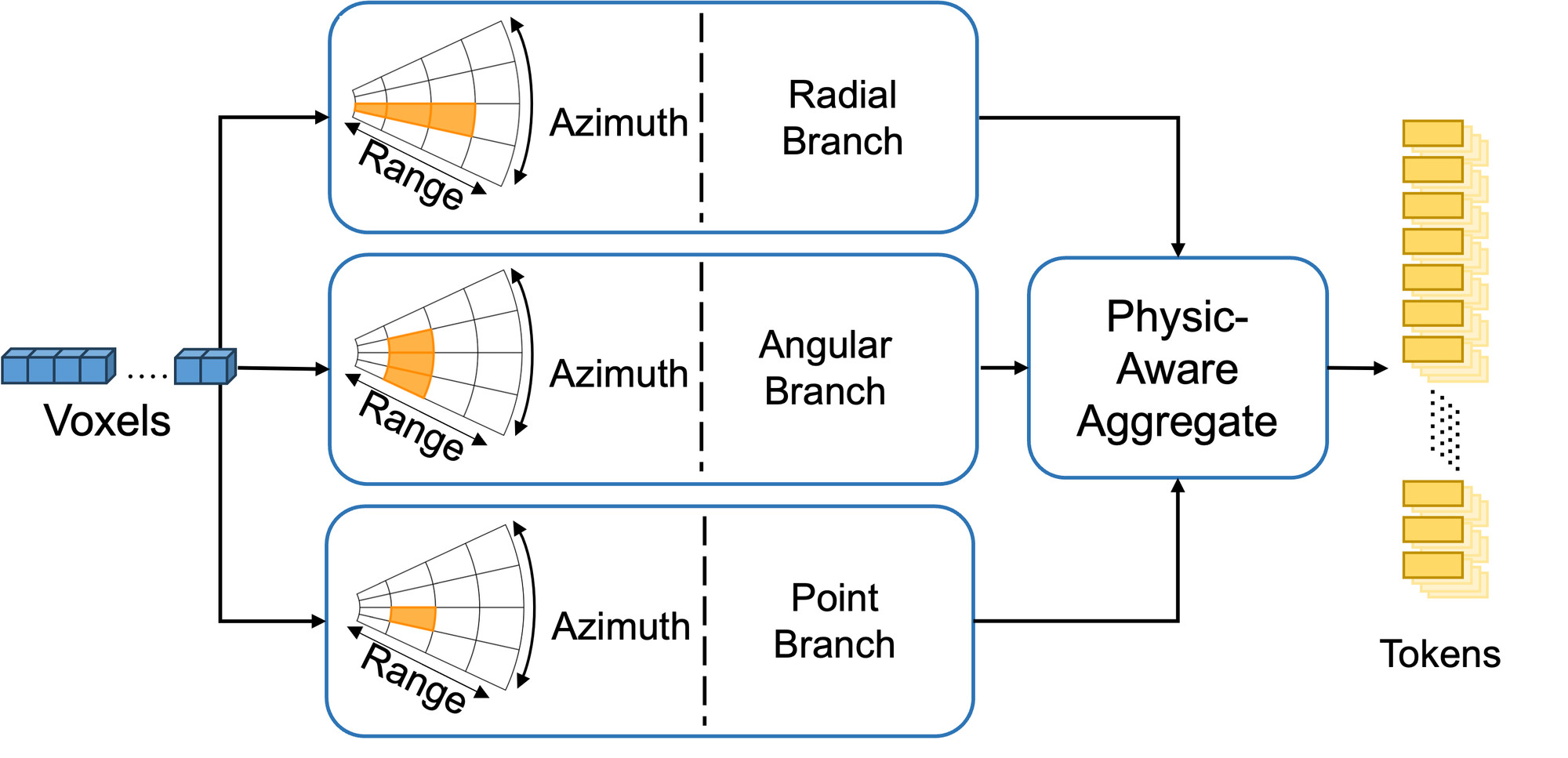}
        \caption{Illustration of physics-aware anisotropic sparse convolution.}
        \label{fig:encoder}
    \end{minipage}
\end{figure}

\section*{Appendix B: Details of \pname's Physical-Aware Encoder}
The point cloud data acquired by an RF sensing system is naturally represented in spherical coordinates. In the spherical coordinate system as illustrated in Figure~\ref{fig:voxel}, the spatial extent covered by a voxel increases with radial distance (\textit{i.e.}, the near-far effect). Conventional 3D convolution operations treat the voxel dimensions as isotropic, which is effective for feature extraction in Cartesian space but ineffective in spherical space.

To address this issue, \pname\ employs sparse convolution across three parallel branches (\textit{i.e.}, radial branch, angular branch, and point branch) for voxel-based feature extraction, as illustrated in Figure~\ref{fig:encoder}. The radial branch captures local continuity along the range axis, while the angular branch captures contextual patterns on the azimuth-elevation plane.
Furthermore, \pname\ incorporates a point-wise residual branch to preserve local voxel responses without neighborhood mixing, thereby stabilizing feature fusion and reducing the risk of over-smoothing. The features encoded by the three branches are then linearly aggregated, followed by positional encoding in both spatial and temporal dimensions. The resulting encoded tokens are subsequently processed by spatial and temporal attention modules for further feature extraction.

\section*{Appendix C: Details of \pname's Heatmap Head}

Given the large number of voxels, the pose estimation network appears to have sparse gradients as the loss function is hard to capture every dimension of the input data.
To address this issue, we adopt a 2.5D pose heatmap 
in the \((r,\theta)\) and \((r,\phi)\) planes as the output of the model, following the principle presented in~\cite{iqbal2018hand}, which provides denser information that can significantly reduce the gradient sparsity. 
To generate the 2.5D pose heatmaps, the conv/deconv head reshapes the target-frame tokens $X_{out}$ into a spatial map and upsamples it using $L$ transposed convolutions to produce the range-azimuth heatmap \(\mathbf{RA}\) and elevation map \(\mathbf{E}\) :
\begin{align}
    &\mathbf{F}_0 = \mathrm{Reshape}(\mathbf{X}_{\text{out}}), \quad
     \mathbf{F}_{l+1} = \sigma(\mathrm{Deconv}(\mathbf{F}_{l})),\\
    & \mathbf{RA} = \mathrm{Conv}_{1\times1}^{\mathrm{RA}}(\mathbf{F}_L), \quad \mathbf{E} = \mathrm{Conv}_{1\times1}^{\mathrm{E}}(\mathbf{F}_L),
\end{align}
where \(\sigma\) is an activation function and \(l = 0,1,\dots, L-1\). 


For joint $j$ in a human pose, its predicted coordinate \(\mathbf{p}_j\) is derived by soft-argmax on heatmap \(\mathbf{RA}_j\) and \(\mathbf{E}_j\). Let $\mathbf{\Omega}$ be the set of its discrete locations on heatmap $\mathbf{RA}$; denote \(\mathbf{RA}_j(v)\) as the heatmap value of its bin index $v \in \mathbf{\Omega}$.
Then, the confidence of joint $j$ is computed by:
\begin{align}
c_j = 1 - \left( - \frac{1}{\log|\mathbf{\mathbf{\Omega}}|}\sum_{u\in \Omega} \mathcal{P}_j(v) \log\left(\mathcal{P}_j(v) \right)\right),
\label{eq:confidence}
\end{align}
where 
$\mathcal{P}_j(v\in \mathbf{\Omega}) = \frac{\mathbf{RA}_{j}(v)}{\sum_{v'\in \mathbf{\Omega}} \mathbf{RA}_{j}(v')+\epsilon}
$, and 
\(\epsilon\) is a small value to avoid zero-division.

\section*{Appendix D: Details of \pname's Network Implementation}

\pname voxelizes each input frame into 20,480 voxels, and five consecutive radio frames are fed into the network for pose estimation. 
Three layers of physics-aware sparse convolution are employed to encode each voxel into a 512-dimensional token. 
An attention-based downsampling layer then reduces the number of tokens from 20,480 to 2,048. 
Six layers of four-head Transformers are applied to model spatial dependencies.
The spatial tokens from the five frames are separated into target-frame tokens and non-target-frame tokens. 
For each non-target frame, the tokens are further downsampled to 576 tokens and concatenated with the target-frame tokens. 
Temporal dependencies are captured using two layers of four-head Transformers.
The heatmap head consists of three convolutional layers followed by three deconvolutional layers to decode the joint heatmaps. 

For IK implementation, we use the 6th joint as the root joint and a three-layer GCN with 512 hidden dimensions as the IK head. 
Recall that \pname has two key components in its network: 
\emph{pose estimator} and \emph{GCN-based IK}.
When training the pose estimator, the loss weights are set to 100 for $\mathcal{L}_{\mathrm{map}}$, 
5 for $\mathcal{L}_{\mathrm{xyz}}$, 
0.3 for $\mathcal{L}_{\mathrm{ang}}$, 
and 3.5 for $\mathcal{L}_{\mathrm{len}}$. 
When training the GCN for IK, the loss weights are adjusted to 0.2 for $\mathcal{L}_{\mathrm{ang}}$, 
0.5 for $\mathcal{L}_{\mathrm{len}}$, 
and 3 for $\mathcal{L}_{\mathrm{xyz}}$. 
The weight of root loss $\mathcal{L}_{\mathrm{root}}$ is set to 0.5.

\begin{figure}[h] 
    \centering
    \includegraphics[width=\linewidth]{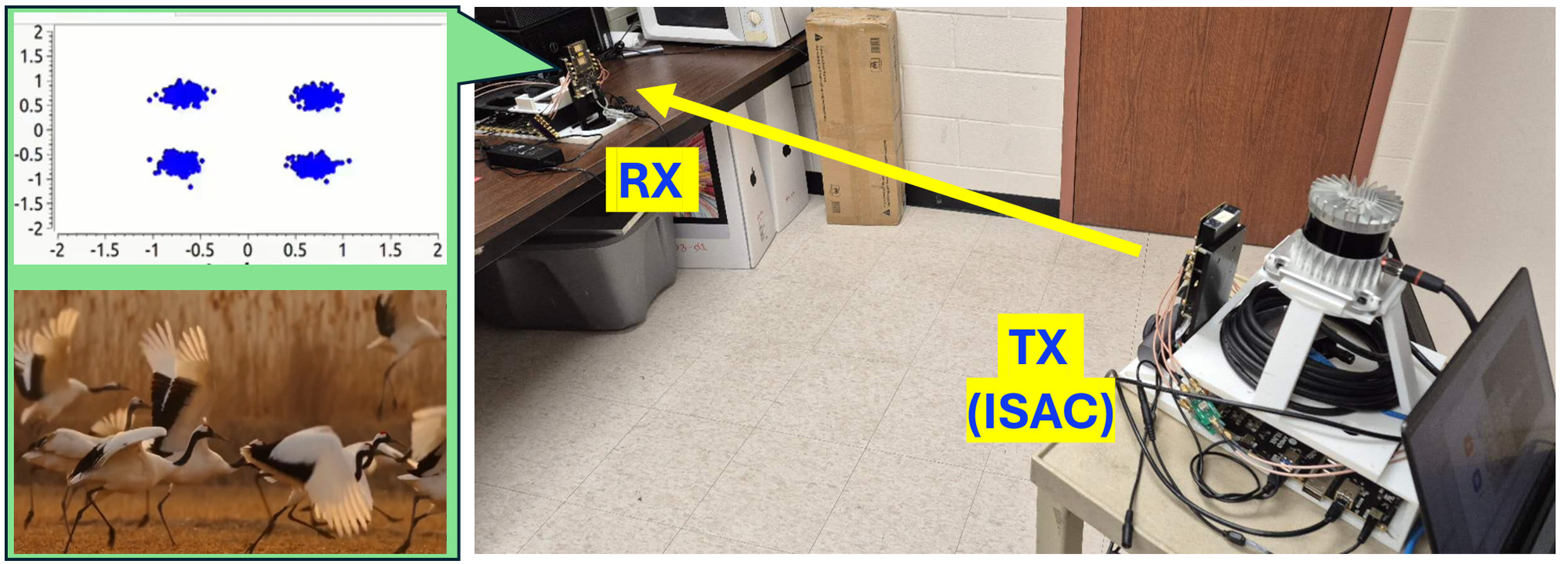}
    \caption{Illustration of \pname performing sensing and communication simultaneously.}
    \label{fig:isac22}
    \vspace{-0.1in}
\end{figure}

\begin{figure*}[t]
\vspace{-0.05in}
    \centering
    \begin{subfigure}[b]{0.3\linewidth}
        \centering
        \includegraphics[width=\textwidth]{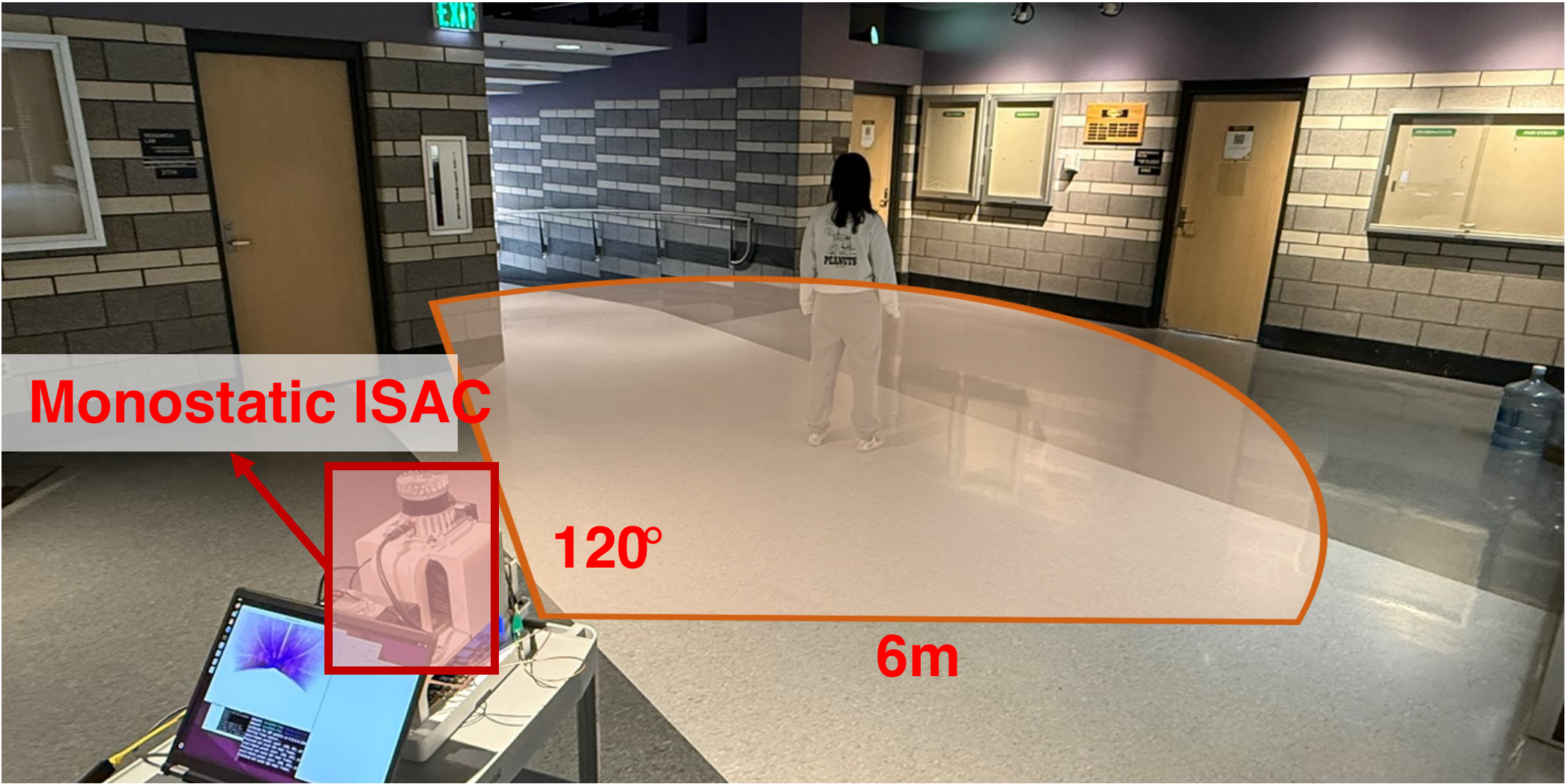}
        \caption{Set-up of \pname}
        \label{fig:setup-isac}
    \end{subfigure}
    \hfill
    \begin{subfigure}[b]{0.3\linewidth}
        \centering
        \includegraphics[width=\textwidth]{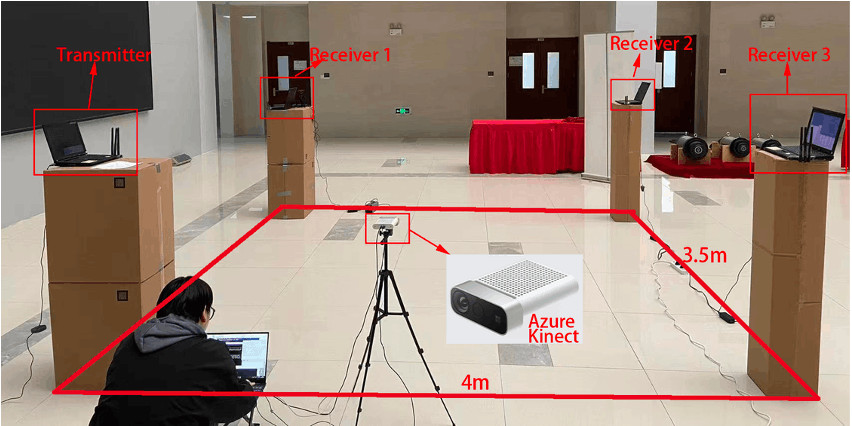}
        \caption{Set-up of Wi-Fi in~\cite{yan2024person}}
        \label{fig:setup-wifi}
    \end{subfigure}
    \hfill
    \begin{subfigure}[b]{0.3\linewidth}
        \centering
        \includegraphics[width=\textwidth]{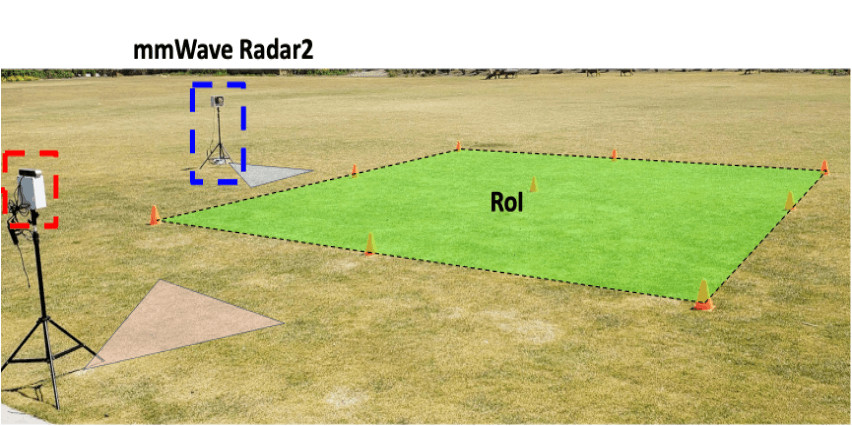}
        \caption{Set-up of MVDoppler~\cite{choi2025mvdoppler}}
        \label{fig:setup-mvdoppler}
    \end{subfigure}
    \vspace{-0.1in}
    \caption{Experimental setups comparison of \pname, Wi-Fi and mmWave radar.}
    \label{fig:comparison-of-setups}
    \vspace{-0.0in}
\end{figure*}

\section*{Appendix E: Data Collection and Overall System Setup}

\pname is capable of performing sensing and communication simultaneously. 
Figure~\ref{fig:cir1} illustrates how OFDM signals are utilized for both sensing and communication at the same time. 
During data collection, \pname also transmits data packets to another device. 
Since the communication data path remains unchanged, the communication functionality and performance are preserved.

Figure~\ref{fig:isac22} shows an example where \pname sends data packets to a receiver while performing sensing. 
In this experiment, a video was streamed from \pname to the receiver. 
Table~\ref{tab:system_params} summarizes the communication and sensing parameters. 
The carrier frequency is 60~GHz. 
The bandwidth is 1.2288 GHz and the sampling rate is 1.2288 samples per second.
The OFDM modulation uses 1024 subcarriers.
Among the 1024 subcarriers, 900 are used for data transmission. 
The transmission power is 20~dBm. 
The field of view is 120$^\circ$ for horizon and 60$^\circ$ for elevation. 

Figure~\ref{fig:comparison-of-setups} shows the system and experimental setup of \pname during our data collection. 
The figure also illustrates the experimental setup of our two baselines as a comparison.
It is noteworthy that \pname needs only one communication device for data collection, while the baselines need two or more communication devices for data collection. 

\begin{table}[!ht]
    \centering
    \caption{The parameters of our monostatic ISAC device.}
    \vspace{-0.08in}
    \renewcommand{\arraystretch}{1.08}
    \setlength{\tabcolsep}{2.0pt} 
    \scriptsize
    \begin{tabular}{|l@{\hskip 3pt}l|l@{\hskip 3pt}l|}
    \hline
    \multicolumn{2}{|c|}{\textbf{Hardware parameters}} &
    \multicolumn{2}{c|}{\textbf{Communication parameters}} \\
    \hline
    Sampling rate & 1.2288~GSPS & Waveform & OFDM \\
    Number of Tx antennas & 16 & FFT points & 1024 \\
    Number of Rx antennas & 16 & Valid subcarriers & 900 \\
    Center frequency & 60~GHz & Cyclic prefix length & 276 \\
    Transmission power & 20~dBm & OFDM symbol duration & 1.057~$\mu$s \\
     &  & OFDM symbols/frame & $16\times16 = 256$ \\
     &  & Supporting protocols & 5G and Wi-Fi \\
    \hline
    \multicolumn{4}{|c|}{\textbf{Sensing parameters}} \\
    \hline
    Detection time/frame & 67~$\mu$s & Frames per second & 10 \\
    Theoretical range & 30~m & Practical range & 10~m \\
    Effective ant. (hori.) & 8 & Effective ant. (elev.) & 4 \\
    Hori. antenna spacing & 0.5$\lambda$ & Elev. antenna spacing & 0.5$\lambda$ \\
    Horizontal FoV & [$-60^\circ$, $60^\circ$] & Elevation FoV & [$-30^\circ$, $30^\circ$] \\
    \hline
    \end{tabular}
    \label{tab:system_params}
\end{table}

\section*{Appendix F: Qualitative Illustration of Estimated Poses}

In this section, we provide additional qualitative examples of poses estimated by \pname to visually demonstrate the estimation accuracy and quality. 
Figure~\ref{fig:unseen_case} presents the visualization results, covering various human activities such as walking, sitting, and more. 
The results demonstrate that \pname accurately reconstructs skeletons that closely match the ground-truth from a depth camera. 
All examples are from previously unseen cases, highlighting the generalization capability of \pname.
A demo video is provided on the project website.

\begin{figure}[H]
    \centering
    \includegraphics[width=0.99\linewidth]{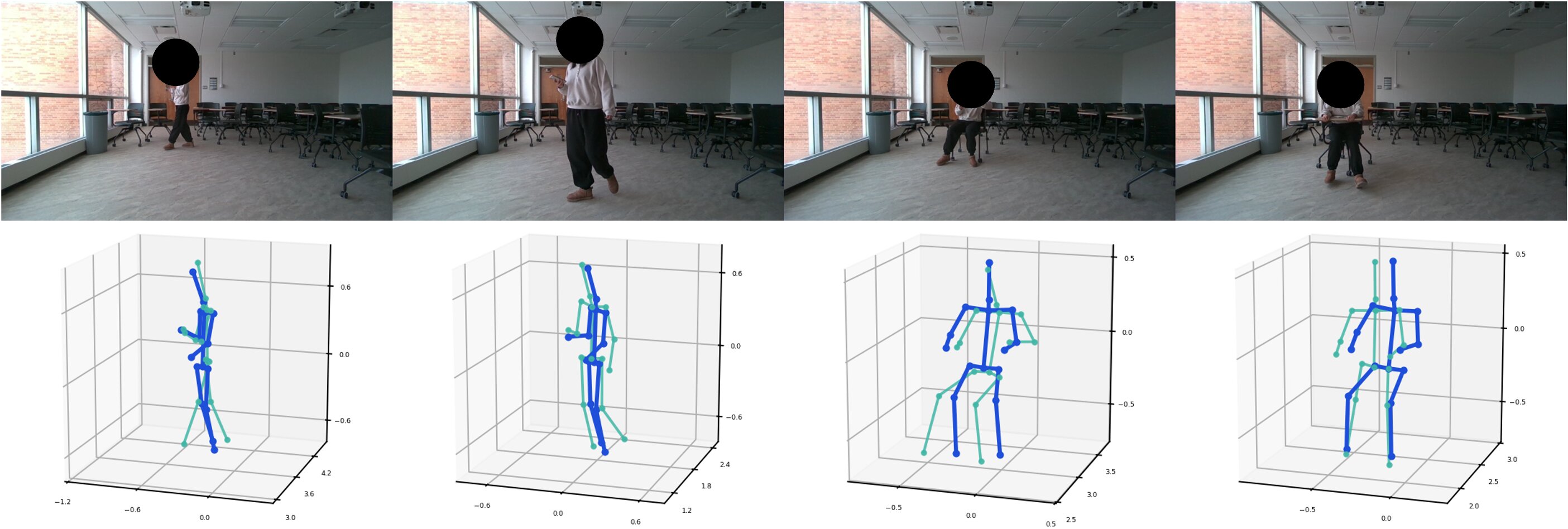}\\
    \includegraphics[width=0.99\linewidth]{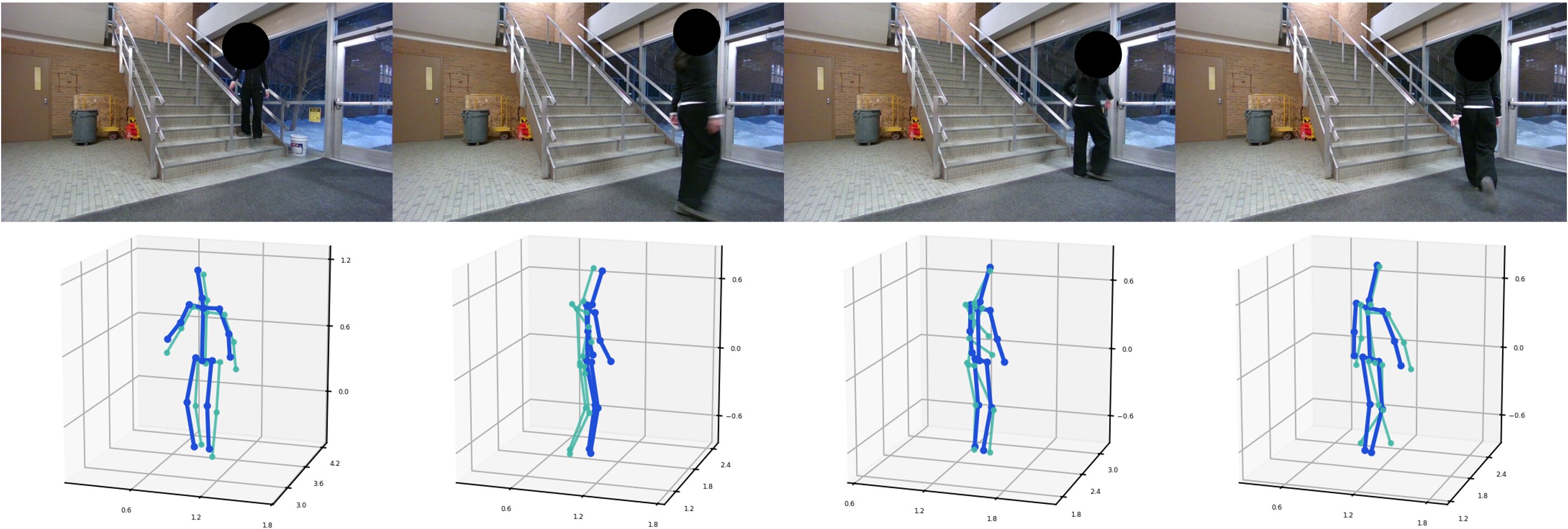}\\
    \includegraphics[width=0.99\linewidth]{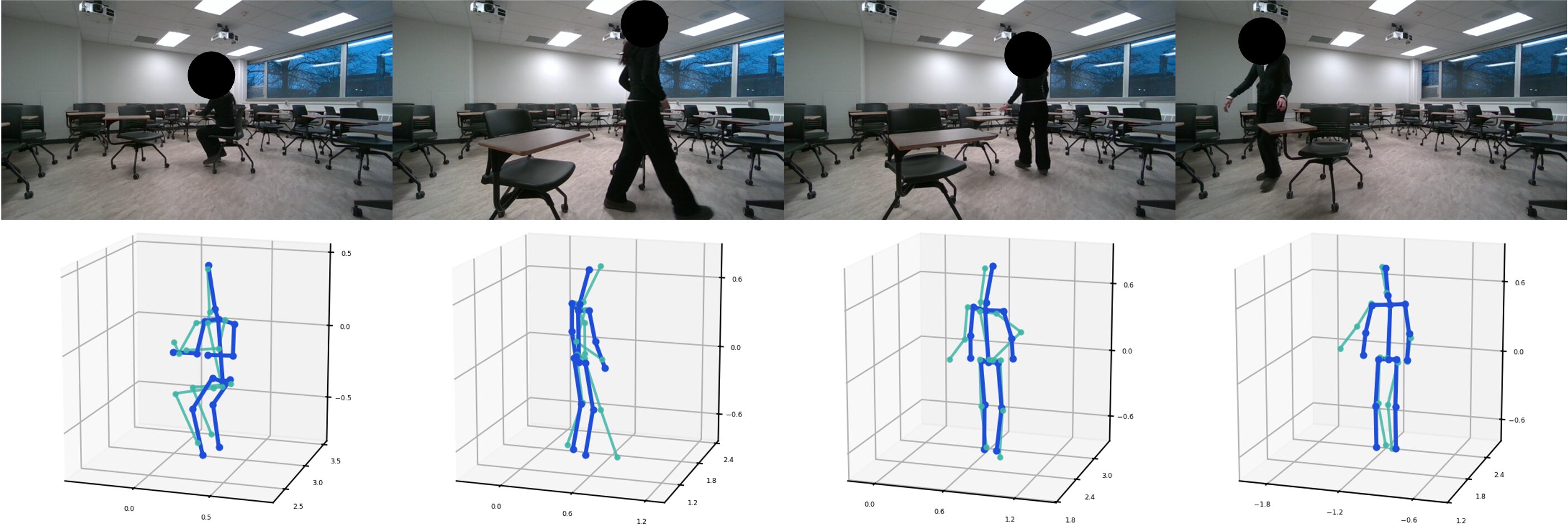}\\
    \includegraphics[width=0.99\linewidth]{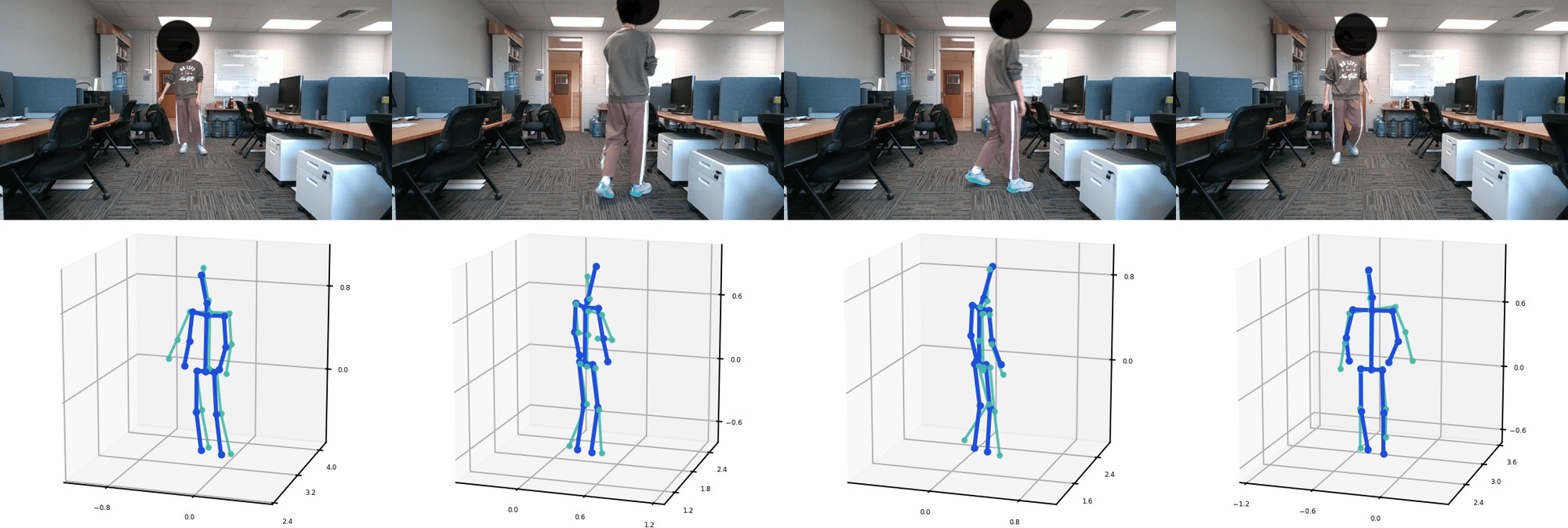}\\
    \caption{Qualitative performance of \pname in \textbf{unseen} scenes.
    Green pose represents ground-truth annotations from depth camera and blue one represents \pname's estimation results.}
    \label{fig:unseen_case}
\end{figure}

\vspace{-0.2in}
\section*{Appendix G: Qualitative Illustration of Robustness Studies}

Compared to vision-based sensors such as cameras and LiDAR, a key advantage of \pname is its robustness to lighting conditions and occlusions. 
In this section, we present additional qualitative results highlighting these advantages. 
Figure~\ref{fig:exp_setting} shows our experimental setup. 
In one scenario, \pname was completely covered with materials such as cloth and foam, while the depth camera remained unobstructed. 
In another scenario, \pname operated in a poorly lit room where the depth camera failed to produce human skeleton.

\begin{figure}[h]
    \centering

    \begin{subfigure}{0.22\linewidth}        
    \includegraphics[width=\linewidth]{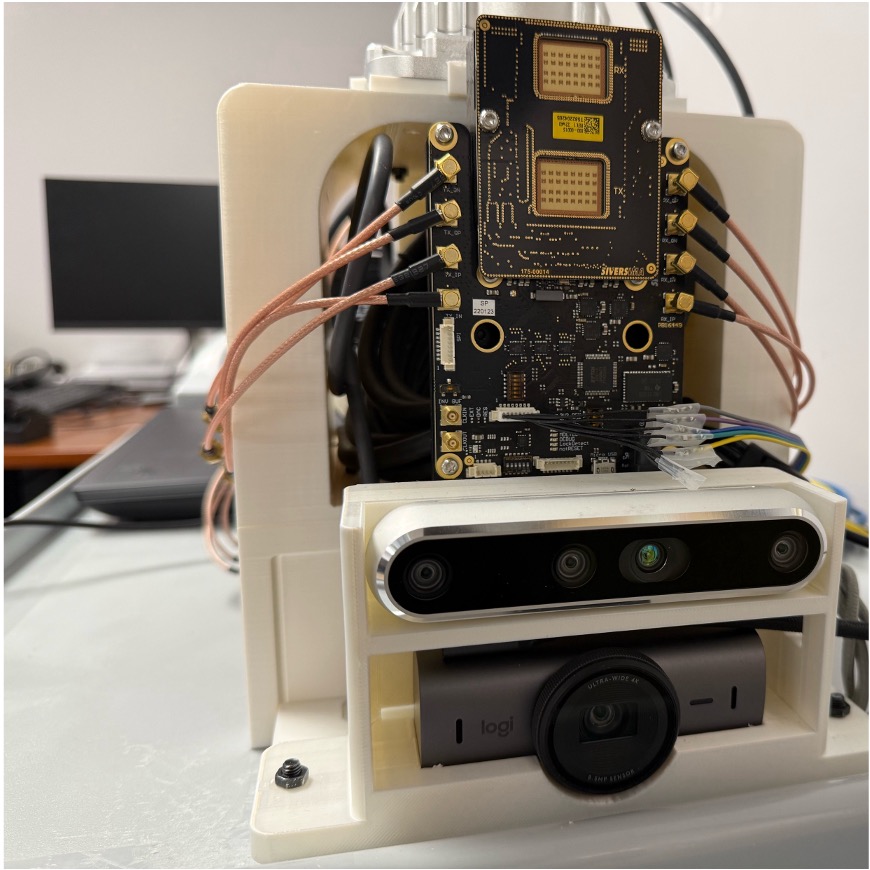}
    \caption{Not Occluded}
    \label{fig:exp_none}
    \end{subfigure}
    \hfill
    \begin{subfigure}{0.22\linewidth}        
    \includegraphics[width=\linewidth]{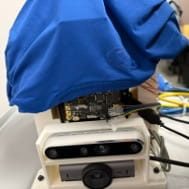}
    \caption{Cloth Occluded}
    \label{fig:exp_cloth}
    \end{subfigure}
    \hfill
    \begin{subfigure}{0.22\linewidth}        
    \includegraphics[width=\linewidth]{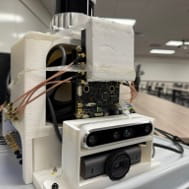}
    \caption{Foam Occluded}
    \label{fig:exp_foam}
    \end{subfigure}
    \hfill
    \begin{subfigure}{0.22\linewidth}        
    \includegraphics[width=\linewidth]{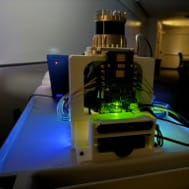}
    \caption{Dim Lighting}
    \label{fig:exp_dim}
    \end{subfigure}\vspace{-0.075in}
    \caption{Experimental setup for different occlusion and lighting conditions.}
    \label{fig:exp_setting}

\end{figure}

\noindent\textbf{Complete Blockage.}
Figure~\ref{fig:occluded_exp} presents several cases where the radio device antennas are completely covered by cloth and foam. In the figures, Green poses represent ground-truth annotations and blue poses represent estimated results.
The results show that \pname can still accurately estimate the pose of the subjects, thanks to the penetration capability of mmWave. 

\begin{figure}[ht]
    \centering

    \begin{subfigure}{0.9\linewidth}        
    \includegraphics[width=\linewidth]{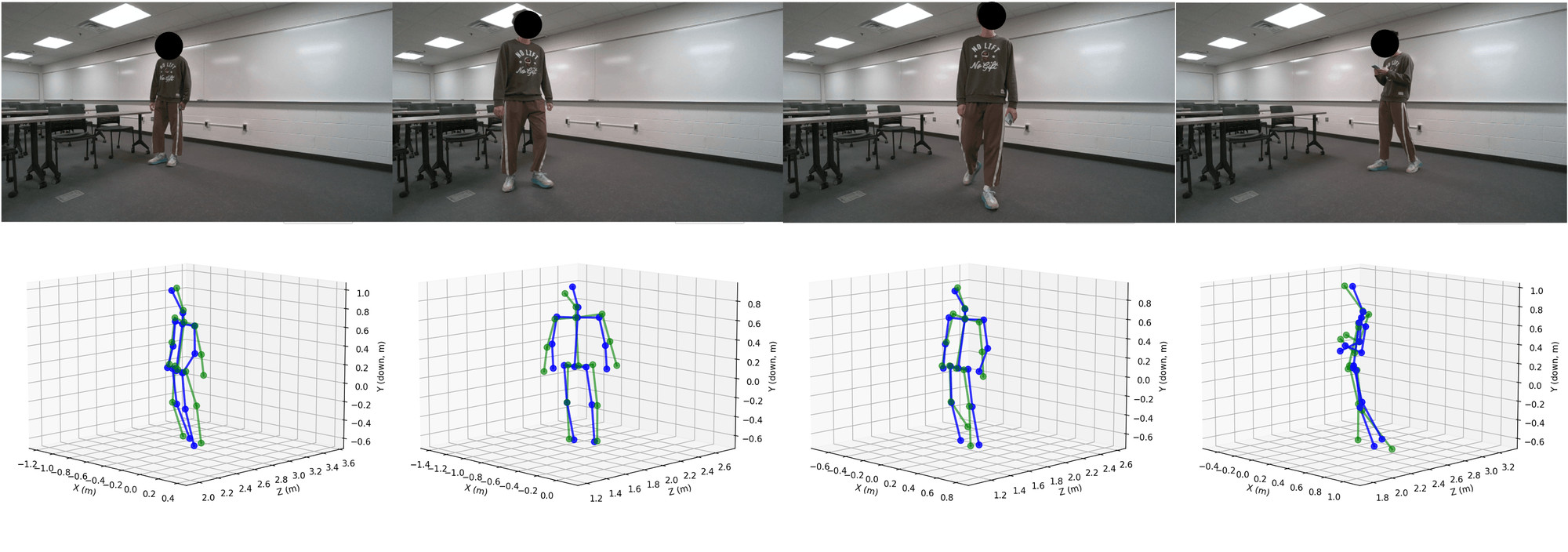}
    \caption{Cloth Occluded.}
    \label{fig:cloth}
    \end{subfigure}\\
    \begin{subfigure}{0.9\linewidth}        
    \includegraphics[width=\linewidth]{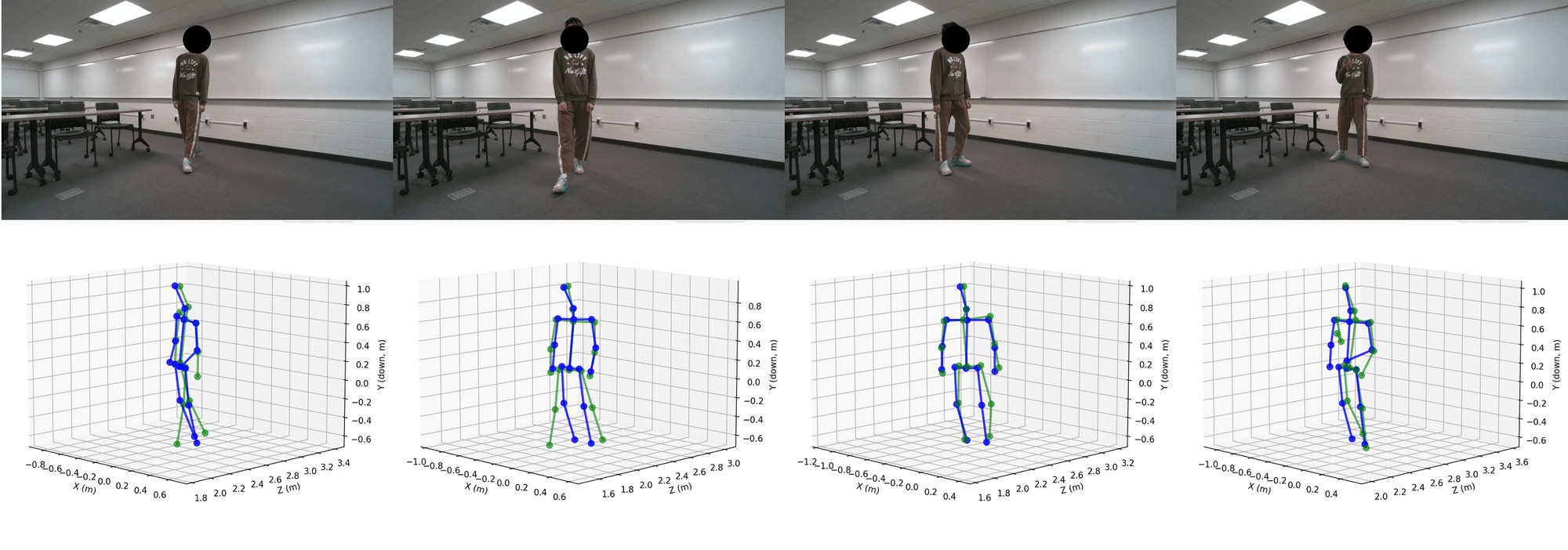}
    \caption{Foam Occluded.}
    \label{fig:foam}
    \end{subfigure}\vspace{-0.075in}
    \caption{Qualitative performance of \pname occluded by cloth and foam. }
    \label{fig:occluded_exp}
\end{figure}

\noindent\textbf{Poor-Lighting Condition.} 
Figure~\ref{fig:dim_results} presents the results under poor-lighting conditions. 
For illustration purpose, we manually enhanced he brightness and contrast of the RGB images to make the scenes visible. 
These adjusted RGB images serve as a reference for qualitative comparison.
In this scenario, the depth camera fails to operate. 
In contrast, \pname is capable of estimating the pose skeleton of the subjects. 
This is because \pname does not rely on lighting illumination. 
We note that, since depth camera fails, there is no ground-truth annotations in the figure.

\begin{figure}[h]
    \centering
    \includegraphics[width=0.99\linewidth]{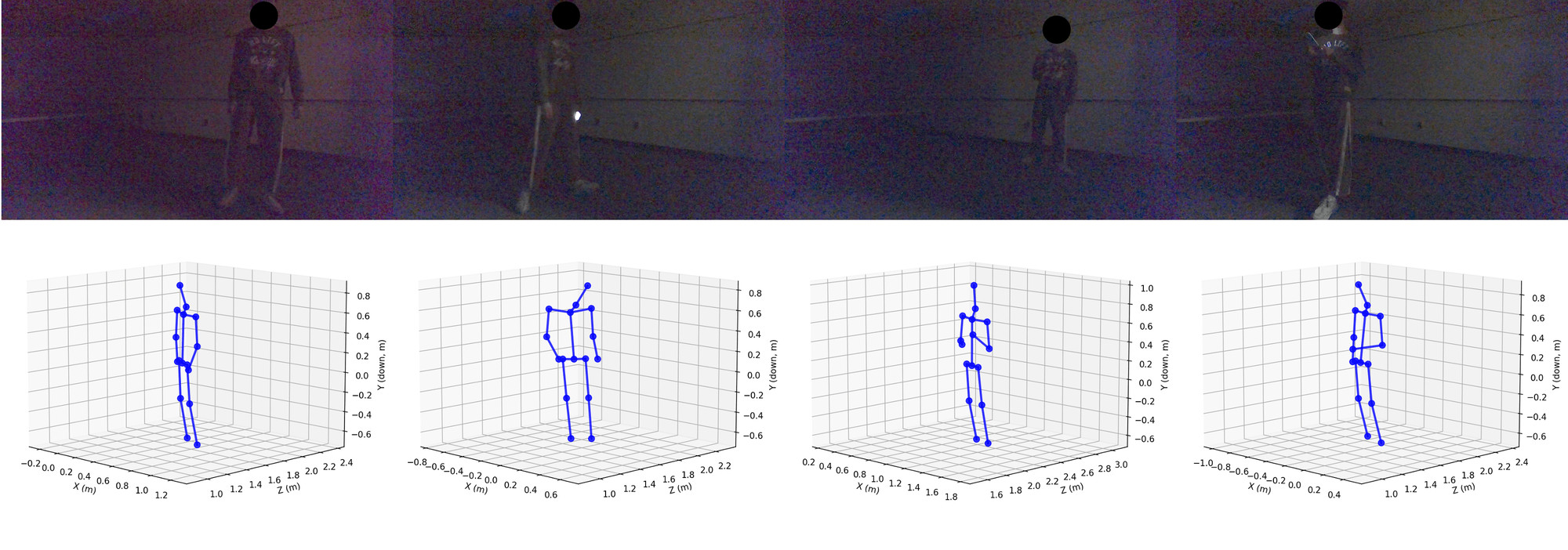}\vspace{-0.1in}
    \caption{Qualitative performance of \pname under low-lighting conditions where depth camera does not work. The brightness and contrast of RGB images have been adjusted to make it visible. }
    \label{fig:dim_results}
\end{figure}


\section*{Appendix H: Qualitative Illustration of Failure Cases}

In this section, we present some failure cases of poses estimated by \pname. 
Figure~\ref{fig:failed-case} illustrates several such instances. 
Most failure cases occur during human activities such as sitting and squatting, as shown in the first and third images, where joints become closely clustered. 
As a result, RF signals are prone to scattering-center merging and self-occlusion, which are difficult for a single-view sensor to resolve. 
Failures also occur when humans are close to background objects such as wall corners or metal railings. 
These objects have large radar cross-sections, causing strong reflections that mask the human body signals. 
Moreover, multipath effects exacerbate false detections and increase localization errors, particularly when subjects are near metallic surfaces.

\begin{figure}[!h]
    \centering

    \includegraphics[width=0.99\linewidth]{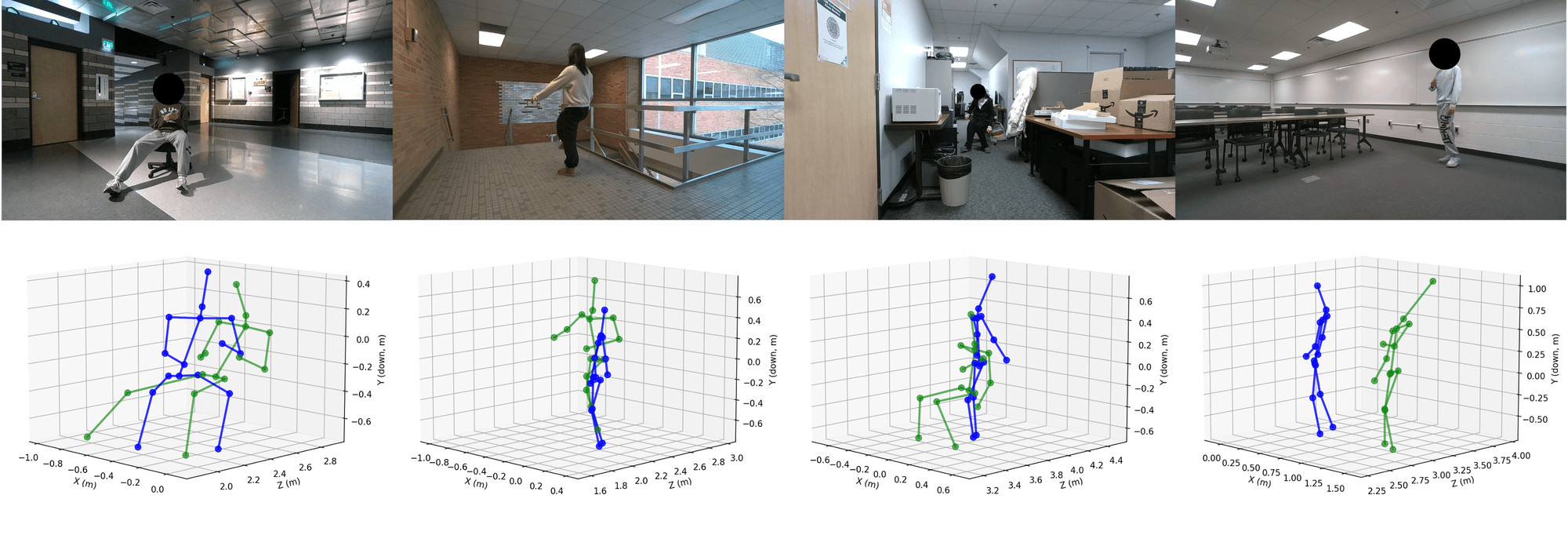}\vspace{-0.1in}
    \caption{Illustration of \pname's failure cases in unseen scenes.}
    \label{fig:failed-case}

\end{figure}

\section*{Appendix I: Additional Experimental Results}

In this section, we present additional experimental results to analyze the performance of \pname. 




\noindent\textbf{Effectiveness of Heatmap Head.}
Instead of performing direct regression, \pname employs a heatmap head to predict a 2.5D pose representation for supervised training. 
To evaluate the effectiveness of the heatmap head, we plot the training loss over iterations. 
Figure~\ref{fig:loss_curve_compare_smoothed} presents the experimental results. 
The results show that, when using direct regression, the training of \pname fails to converge due to sparse and unstable gradients. 
In contrast, when the proposed 2.5D heatmap-based loss is used, the training converges fast, demonstrating the effectiveness of the heatmap head.


\begin{figure}
    \centering
    \includegraphics[width=0.7 \linewidth]{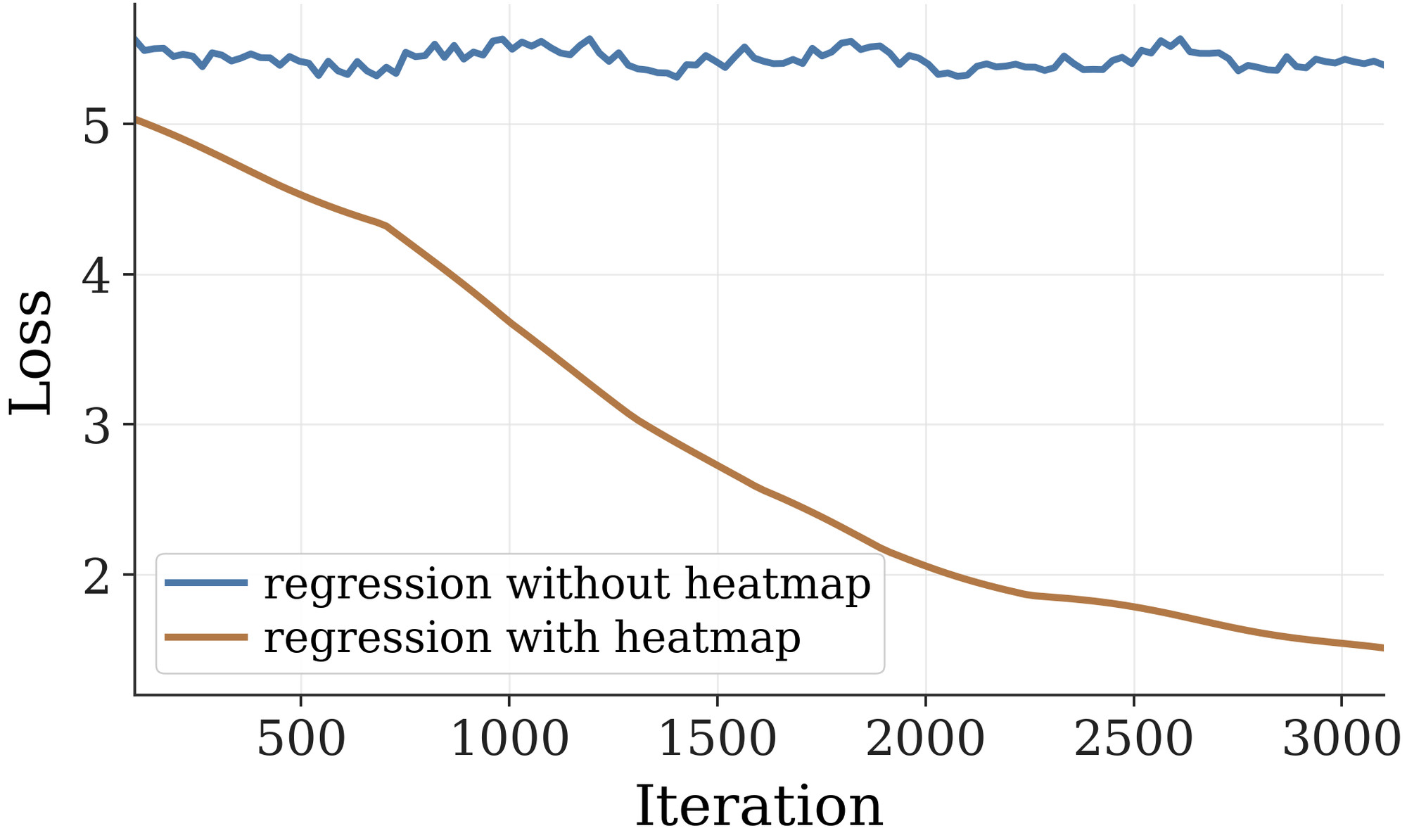}\vspace{-0.1in}
\caption{Training convergence comparison between direct regression and heatmap head.}
\label{fig:loss_curve_compare_smoothed}
\end{figure}

\noindent\textbf{Ablation Study of PASC.}
\pname employs Physics-Aware Sparse Convolution (PASC) as its encoder to mitigate the near-far effect. 
We conducted an ablation study to evaluate its effectiveness. 
We perform ablation studies to evaluate its effectiveness. 
Table~\ref{tab:ablation-pasc} presents the results. 
Without PASC, \pname achieves an MPJPE of 17.1\,cm, whereas incorporating PASC reduces the MPJPE to 16.2\,cm.
Although the gain is small, PASC is justified because it does not increase computational cost compared to conventional 3D convolution operations.

\begin{table}[]
    \centering
    \footnotesize 
        \caption{Effectiveness of PASC.}
        \vspace{-0.1in}
        \label{tab:ablation-pasc}
        \begin{tabular}{lcc}
        \toprule
         PASC  \ & \  MPJPE (cm)\(\downarrow\) \  &\ \(\text{PCK}_{10}\uparrow\) \\
        \midrule
          w/o &  17.1 &  29.9\\
          w  &   \textbf{16.2} &  \textbf{36.3}\\
        \bottomrule
        \end{tabular}
\end{table}

\noindent\textbf{Ablation Study of TDTA.}
Determined by the spectrum availability, the ranging resolution of \pname is 12.2\,cm, posing challenges in capturing reflections from small joints like wrists and ankles.
Therefore, temporal attention becomes essential to improve the pose estimation accuracy. 
The target-driven temporal attention (TDTA) module in our model was designed to address this challenge.
We now perform ablation studies to evaluate its effectiveness.

Table~\ref{tab:TDTA} presents the pose estimation accuracy of \pname using different temporal attention mechanisms.
It includes two baselines: (i) no temporal attention and (ii) standard temporal attention for information aggregation. 
The results show that standard temporal attention does not improve performance and sometimes even degrades it, likely due to the introduction of additional noise from multiple frames. 
In contrast, \pname's TDTA approach (temporal attention with a target-frame query) yields a significant improvement for pose estimation accuracy.


\begin{table}[!t]
\centering
\scriptsize
\captionof{table}{Effectiveness of target-driven temporal attention (TDTA).}
\label{tab:TDTA}
\setlength{\tabcolsep}{3pt}
\begin{tabular}{p{2in}cc}
\toprule
Temporal Attention Mechanism & MPJPE (cm)\(\downarrow\) & $\text{PCK}_{10}\uparrow$ \\
\midrule
No Temporal Attention & 20.8 & 6.1 \\

Standard Temporal Attention & 22.0 & 7.5 \\

\textbf{Temporal Attention with Target Frame Query} & \textbf{16.2} & \textbf{36.3} \\
\bottomrule
\end{tabular}
\vspace{-0.1in}
\end{table}

\begin{figure*}[!t]
    \centering
    \includegraphics[width=0.9\linewidth]{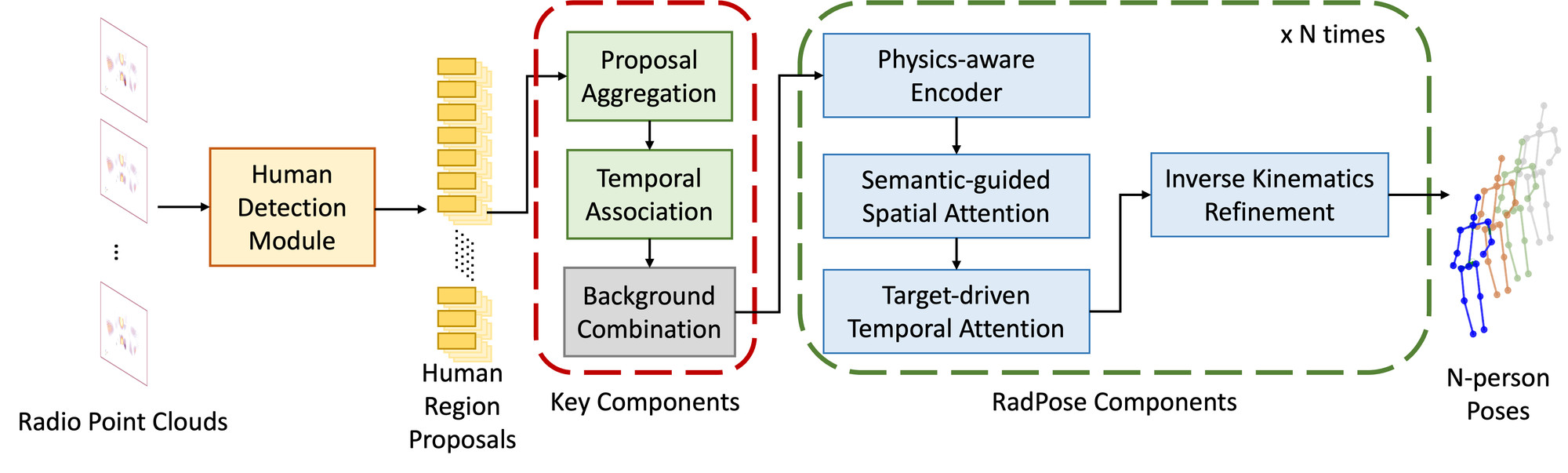}\vspace{-0.1in}
    \caption{A possible conceptual framework that extends \pname to multi-person pose estimation.}
    \label{fig:multi-person}
\end{figure*}

\noindent\textbf{Impact of Voxelization.}
The voxel size introduces a trade-off between accuracy and computational cost. 
On one hand, using a larger number of voxels can potentially improve performance; on the other hand, it significantly increases GPU memory consumption.
Table~\ref{tab:para_voxels} reports the measured MPJPE and the required GPU memory when \pname uses different voxel configurations. 
Increasing the number of voxels from 20,480 to 30,720 improves MPJPE by only 0.3\,cm. 
However, this configuration requires approximately 1.5\,GB of additional GPU memory per input sample. 
Since this marginal accuracy gain does not justify the increased memory usage, we adopt 20,480 voxels as a balanced configuration.

\begin{table}[tbp]
    \centering
    \footnotesize 
    \caption{Comparison of different voxels.}\vspace{-0.1in}
    \label{tab:para_voxels}
    \begin{tabular}{lcc}
        \toprule
        Voxels & Memory (GB) & MPJPE(cm)\(\downarrow\)\\
        \midrule
        10240   & 4.9  & 19.7  \\
        20480   & 6.4  & 16.2 \\
        30720   & 7.9  & 15.9\\
        \bottomrule
    \end{tabular}
\end{table}

\noindent\textbf{Impact of Frame Number.}
RF-derived environmental features are inherently sparse. To mitigate this sparsity, \pname fuses multiple consecutive frames for pose estimation. 
Intuitively, incorporating more frames may improve performance by providing richer temporal information. 
To evaluate the impact of the number of frames, we conducted experiments by increasing the fusion window from three to seven frames. 
Table~\ref{tab:para_frames} presents the results. 
When five frames are used for pose estimation, \pname achieves an MPJPE of 16.2\,cm. However, increasing the number of frames to seven does not further improve accuracy; instead, the MPJPE slightly increases to 16.9\,cm. 
This result suggests that incorporating excessive temporal information may introduce noisy or redundant features that degrade pose estimation performance.

\begin{table}[htbp]
    \centering
    \footnotesize 
    \caption{Comparison of different frame numbers.} 
    \label{tab:para_frames}
    \begin{tabular}{lcc}
        \toprule
        Frames \ \ \  &   Memory (GB)  &\  MPJPE (cm)\(\downarrow\) \\
        \midrule
        3 & 4.5 & 18.6\\
        5 & 6.4 & 16.2\\
        7 & 8.3 & 16.9\\
        \bottomrule
    \end{tabular}
\end{table}

\noindent\textbf{Impact of Spatial Transformer Layers.}
After voxelization and encoding, \pname uses multiple layers of spatial transformers to capture spatial relationships. 
Using more transformer layers can enhance representation capacity, but it also increases GPU memory consumption.
To evaluate this trade-off, we conducted experiments by increasing the number of transformer layers from four to eight. 
Table~\ref{tab:para_layers_transformer} reports the results. 
Adding more layers is hard to further improve MPJPE, but it significantly increases GPU memory usage. 
Considering the trade-off between performance and computational cost, we adopt six layers in the final model configuration.

\begin{table}[htbp]
    \centering
    \footnotesize 
    \caption{Comparison of different layers of transformer.} 
    \label{tab:para_layers_transformer}
    \begin{tabular}{lcc}
        \toprule
        Layers \ \ \   &   Memory (GB)  &\  MPJPE (cm)\(\downarrow\) \\
        \midrule
        4 & 5.8 & 21.4\\
        6 & 6.4 & 16.2\\
        8 & 6.9 & 16.1\\
        \bottomrule
    \end{tabular}        
\end{table}

\noindent\textbf{Impact of Transformer Dimension.}
In addition to the number of transformer layers, the embedding dimension of the transformer also affects the performance of \pname. 
To evaluate its impact, we conducted experiments by increasing the embedding dimension from 128 to 512. 
Table~\ref{tab:para_dims_transformer} reports our results. 
Increasing the dimension from 128 to 256 significantly improves pose estimation accuracy, reducing the MPJPE from 23.6\,cm to 18.0\,cm. 
Further increasing the dimension to 512 yields the best performance, achieving an MPJPE of 16.2\,cm. 
We adopt an embedding dimension of 512 in the final model to balance performance and computational cost.

\begin{table}[htbp]
    \centering 
    \footnotesize 
    \caption{Comparison of different dimensions of transformer.}
    \label{tab:para_dims_transformer}
    \begin{tabular}{lcc}
        \toprule
        Dimensions  &   Memory (GB)  &\  MPJPE (cm)\(\downarrow\) \\
        \midrule
        128 & 2.8 & 23.6 \\
        256 & 4.0 & 18.0\\
        512 & 6.4 & 16.2\\
        \bottomrule
    \end{tabular}
\end{table}

\section*{Appendix J: Computational Cost and Latency}

\pname contains 174.9 million parameters and predicts one pose skeleton from a five-frame input window. We measure latency on preloaded data using CUDA synchronization, excluding storage I/O. In streaming inference, each incoming frame is processed using a rolling five-frame window, resulting in an end-to-end processing latency of 68.1\,ms on a single GPU. Because radio data are acquired at 10\,Hz, corresponding to a 100\,ms sampling interval, this latency supports real-time operation with computation offloaded to an edge GPU.

During inference, \pname uses 6.46\,GB of GPU memory, indicating that \pname can be deployed on commonly available GPU accelerators without the substantially higher memory demands of training. Table~\ref{tab:comp_cost} summarizes the model size, computational cost, resource usage, and latency.

\begin{table}[htbp]
    \centering 
    \footnotesize 
    \caption{Computation cost and latency of \pname inference.}
    \label{tab:comp_cost}
    \begin{tabular}{lc}
        \toprule
        Property  &   Value \\
        \midrule
        Parameters & 174.9M \\
        \# of input frames & 5 \\
        Inference GPU & Nvidia RTX 6000 Blackwell\\
        Streaming latency & 68.1\,ms\\
        Inference memory & 6.46\,GB\\
        \bottomrule
    \end{tabular}
\end{table}

\section*{Appendix K: Extending \pname to Multi-Person Pose Estimation}

Currently, \pname is trained and evaluated in single-person scenarios. Its framework could be extended to multi-person settings by, for example, following the approaches in~\cite{zhao2018rf,10.1145/3745756.3809227}. Specifically, a human-detection module could first identify person-specific regions within the dense radio point clouds and then apply \pname's single-person pose estimator to each detected region. Because \pname reconstructs 3D poses from spatially structured radio point clouds without relying on fixed subject identities, the same pose estimation model can be shared across all detected regions.

Several challenges must be addressed to realize this extension. RF responses from multiple subjects may overlap, while multipath propagation makes it difficult to associate each subject's responses consistently across frames. Simply cropping individual regions and estimating their poses may discard useful contextual information, resulting in spatially misaligned or temporally inconsistent poses. One potential solution, illustrated in Figure~\ref{fig:multi-person}, is to aggregate human-region proposals and associate them across time using learnable modules. Robustly addressing these challenges is beyond the scope of this work; therefore, we leave the implementation and evaluation of multi-person pose estimation to future work.


\end{document}